\documentclass[runningheads]{llncs}

\usepackage{eccv}

\usepackage{eccvabbrv}

\usepackage{graphicx}
\usepackage{booktabs}
\usepackage{multirow}
\usepackage{array}

\usepackage{cuted}
\usepackage{capt-of}

\usepackage[accsupp]{axessibility}  % Improves PDF readability for those with disabilities.

\usepackage{hyperref}

\usepackage{orcidlink}
\usepackage{xcolor}

\begin{document}

% \title{AnyCrowd: Identity-Pose Binding via Gated Decoupled Attention for Arbitrary Multi-Character Animation} 
\title{AnyCrowd: Instance-Isolated Identity-Pose Binding for Arbitrary Multi-Character Animation}

\titlerunning{AnyCrowd}

\author{Zhenyu Xie\inst{1} \and
Ji Xia\inst{1} \and
Michael Christian Kampffmeyer\inst{2} \and
Panwen Hu\inst{1} \and
Zehua Ma\inst{3} \and
Yujian Zheng\inst{1} \and
Jing Wang\inst{3} \and
Zheng Chong\inst{3} \and
xujie zhang\inst{3} \and
Xianhang Cheng\inst{1} \and
Xiaodan Liang\inst{1,3} \and
Hao Li\inst{1}}

\authorrunning{Z.~Xie et al.}

\institute{Mohamed bin Zayed University of Artificial Intelligence, UAE \and
University of Tromsø (UiT) – The Arctic University of  Norway, Norway \and
Shenzhen campus of Sun Yet-sen University, China
% Princeton University, Princeton NJ 08544, USA \and
% Springer Heidelberg, Tiergartenstr.~17, 69121 Heidelberg, Germany
\url{https://xiezhy6.github.io/anycrowd/}
}

\maketitle

{ 
    \centering
    \captionsetup{type=figure}
    \includegraphics[width=1.0\textwidth]{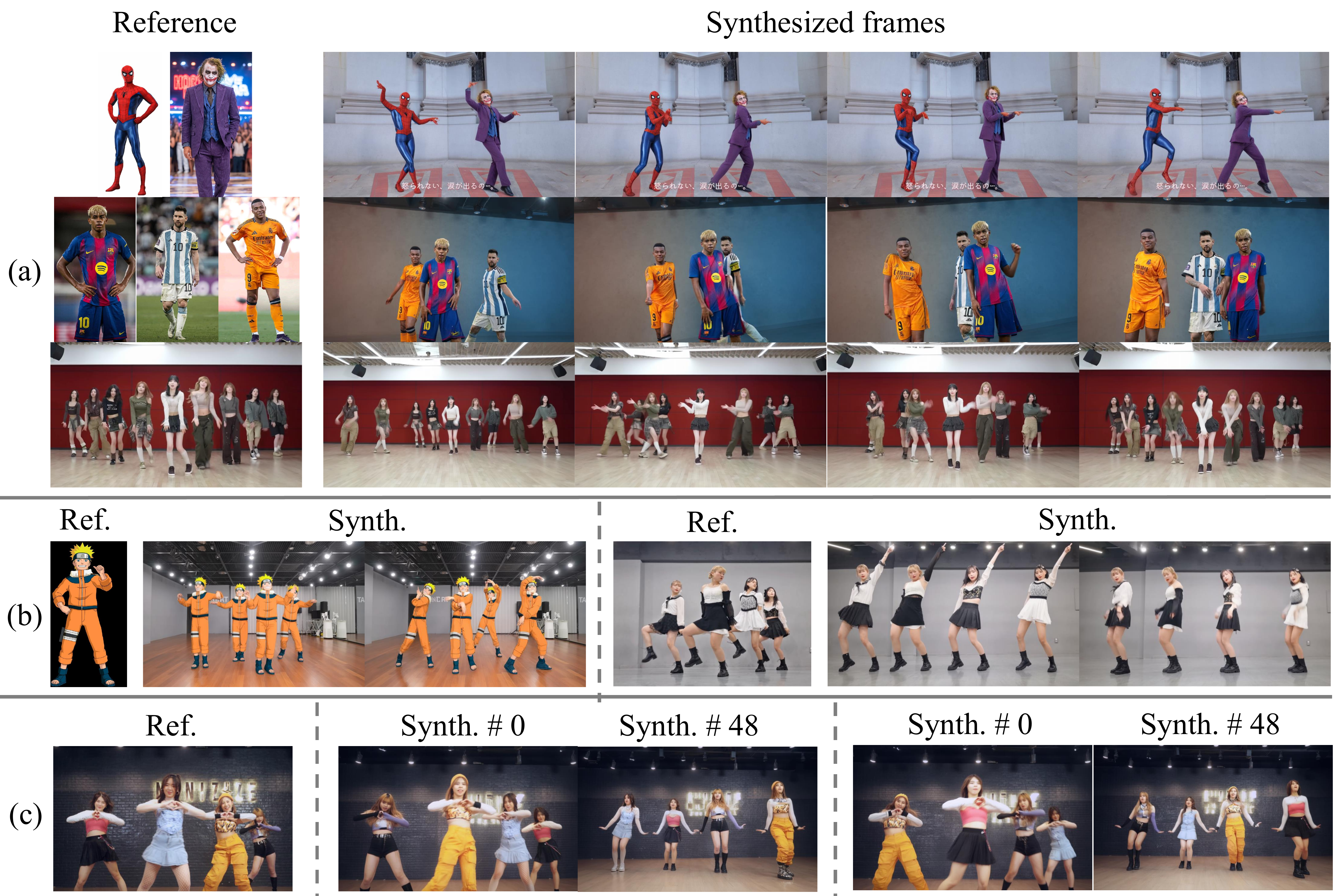}
    \captionof{figure}{We propose AnyCrowd, a versatile framework for multi-character animation, which supports: (a) animation of an arbitrary number of characters sourced from either single or multiple reference images; (b) diverse configurations such as many-to-one (multiple poses driving one identity) or one-to-many (one pose driving multiple identities) generation; and (c) arbitrary assignments between IDs and pose sequences.}
    \label{fig:teaser}
}

\begin{abstract}
Controllable character animation has advanced rapidly in recent years, yet multi-character animation remains underexplored.
As the number of characters grows, multi-character reference encoding becomes more susceptible to latent identity entanglement, resulting in identity bleeding and reduced controllability. Moreover, learning precise and spatio-temporally consistent correspondences between reference identities and driving pose sequences becomes increasingly challenging, often leading to identity-pose mis-binding and inconsistency in generated videos. 
To address these challenges, we propose \textbf{AnyCrowd}, a Diffusion Transformer (DiT)-based video generation framework capable of scaling to an arbitrary number of characters.
Specifically, we first introduce an Instance-Isolated Latent Representation (IILR), which encodes character instances independently prior to DiT processing to prevent latent identity entanglement.
Building on this disentangled representation, we further propose Tri-Stage Decoupled Attention (TSDA) to bind identities to driving poses by decomposing self-attention into: (i) instance-aware foreground attention, (ii) background-centric interaction, and (iii) global foreground-background coordination. 
Furthermore, to mitigate token ambiguity in overlapping regions, an Adaptive Gated Fusion (AGF) module is integrated within TSDA to predict identity-aware weights, effectively fusing competing token groups into identity-consistent representations.
To validate effectiveness and scalability, we curate Multi-Character-Dancing-7K (MCD-7K), containing 7,384 clips ($\sim$31 hours) of 2--6 performers, and establish a held-out benchmark, MCD-300, featuring 2--9 characters per clip. Extensive experiments show that AnyCrowd outperforms state-of-the-art single- and multi-character baselines, with ablations confirming each component's contribution. Notably, AnyCrowd generalizes zero-shot to unseen crowd densities and supports flexible identity-motion recasting.

\end{abstract}

\section{Introduction}
\label{sec:intro}

Benefiting from advances in high-capacity generative models~\cite{rombach2022high,peebles2023scalable,hong2022cogvideo,yang2024cogvideox,kong2024hunyuanvideo,wu2025hunyuanvideo,wan2025wan}, controllable character animation, which aims to synthesize realistic human videos from a reference image and a driving pose sequence, has progressed rapidly in recent years~\cite{hu2024animate,tan2024animate,hu2025animate,wang2025unianimate,jiang2025vace,tu2025stableanimator,cheng2025wan,shao2025interspatial,ding2025mtvcrafter,zhang2024mimicmotion,zhou2025realisdance,zhang2025flexiact}. 
However, existing methods largely focus on \emph{single-character} animation. 
% In contrast, many real-world applications are inherently \emph{Multi-Character Animation (MCdA)} tasks (e.g., group choreography and multi-actor previsualization), 
In contrast, many real-world applications are inherently \emph{Multi-Character Animation (MCA)} tasks, such as group choreography, multi-actor previsualization, and synthetic data generation for Embodied AI (e.g., multi-agent interactions and social navigation). 
Yet, MCA remains comparatively underexplored due to the inherent complexity introduced by multiple interacting identities.

\begin{figure}[tb]
  \centering
  \includegraphics[width=\textwidth]{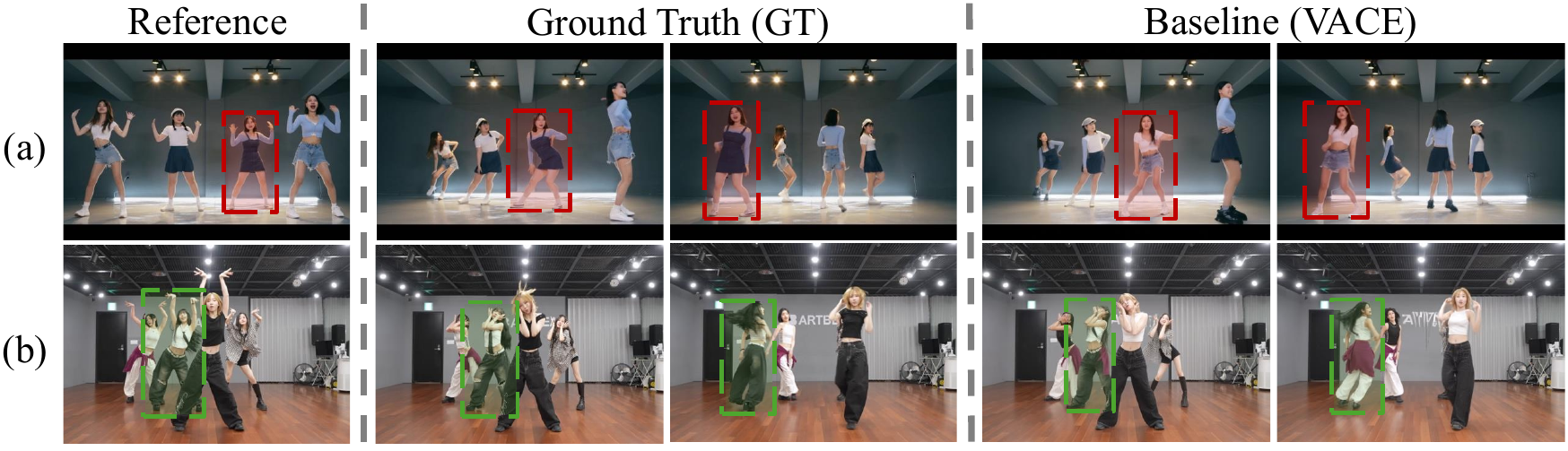}
  \caption{Failure cases of baseline method~\cite{jiang2025vace} in multi-character scenarios. (a) Identity-Pose Mis-Binding (\textcolor{red}{red}): identity swap. (b) Identity Entanglement (\textcolor{green}{green}): appearance blending. Boxes track specific identities across Reference, GT, and generated frames.}
  \label{fig:singleani_issues}
\end{figure}

In practice, naively extending existing animation models~\cite{zhou2025realisdance,wang2025unianimate,jiang2025vace,wan2025wan} to multi-character scenarios often leads to severe artifacts and degraded controllability. Specifically, these methods frequently suffer from \emph{Identity-Pose Mis-Binding}. As illustrated in Fig.~\ref{fig:singleani_issues} (a), the generated character in the red box erroneously adopts the appearance of another individual while following the assigned pose, failing to maintain the Ground Truth (GT) correspondences. This mis-binding stems from the inherent spatial stochasticity in multi-character scenes. Existing frameworks typically employ global self-attention to learn correspondences between a multi-identity reference image and a multi-subject pose sequence. However, the spatial ordering of identities in the reference image often lacks a deterministic alignment with their positions in the driving frames. Without explicit identity-aware priors, the attention mechanism struggles to find correct matches among numerous possibilities. This inevitably leads to "attention drift", where the features of a specific identity are erroneously coupled with an unrelated motion signal, resulting in identity swaps.
Furthermore, \emph{Identity Entanglement} remains a pervasive challenge, becoming particularly pronounced during interactions. As illustrated in Fig.~\ref{fig:singleani_issues} (b), the characters in the green boxes exhibit a blending of visual features with adjacent identities—a degradation that intensifies after spatial transitions, such as an individual in the final frame erroneously incorporates specific visual traits (e.g., trousers texture) from a neighbor. This entanglement stems from the holistic encoding of multi-character reference images. When standard encoders (e.g., VAE) downsample a dense multi-subject image into a compact latent space, a single latent token often encompasses features from multiple adjacent individuals. This leads to latent feature conflation at the source, where distinct identity attributes are irreversibly mixed. Consequently, the model fails to maintain clean, identity-specific representations, resulting in the observed attribute leakage and visual artifacts.

To this end, we propose \textbf{AnyCrowd}, a DiT-based video generation framework designed to overcome these fundamental bottlenecks through instance-level decoupling and precise identity-pose binding.
Specifically, AnyCrowd first presents an Instance-Isolated Latent Representation (IILR) to eliminate identity entanglement. By randomly reassigning and encoding each character individually, IILR disentangles identities from their original spatial context and generates multiple identity-aware reference tokens free from inter-character interference.
Building on these isolated tokens, AnyCrowd utilizes Tri-Stage Decoupled Attention (TSDA) to explicitly bind each identity feature to its corresponding motion track, ensuring that appearances are correctly assigned to the intended characters. Instead of employing the standard global self-attention found in vanilla DiT blocks, TSDA reconfigures the self-attention into three structured stages: (i) instance-aware foreground attention, which extracts tokens of a specific ID from both reference and video frames to compute per-ID self-attention, thereby enforcing an exclusive identity-pose binding; (ii) background-centric interaction, which performs self-attention among background tokens from both sources to maintain environmental consistency; and (iii) global foreground-background coordination, which conducts self-attention solely on the video tokens to facilitate spatial interactions and overall structural coherence.
Crucially, to ensure generalization during inference, AnyCrowd is trained using bounding box masks rather than precise tracking masks to extract character-specific video tokens, thereby preventing the model from exploiting identity shortcuts. While this strategy effectively avoids information leakage, it inevitably introduces spatial overlaps between the extracted tokens of different identities and the background. To resolve these conflicts, AnyCrowd integrates an Adaptive Gated Fusion (AGF) module within TSDA. For tokens in overlapping regions, AGF predicts individual fusion weights for each contributing category (e.g., distinct identities and the background); these weights are then applied to their respective tokens to perform a weighted integration. By doing so, AGF ensures that each category maintains its correct and distinct information even within shared latent regions.

To evaluate the efficacy of AnyCrowd, we train our model on Multi-Character-Dancing-7K (MCD-7K)—more than 7,000 clips ($\sim$31h) with 2--6 characters—and evaluate it on the disjoint MCD-300 dataset, the first benchmark to feature dense scenarios ranging from 2 to 9 characters per clip. Experimental results demonstrate that AnyCrowd consistently outperforms existing single- and multi-character baselines in both identity consistency and motion fidelity. Moreover, AnyCrowd supports versatile applications such as arbitrary character animation and identity-motion recasting (as shown in Fig.~\ref{fig:teaser}).
Our main contributions can be summarized as follows: 
(1) we propose AnyCrowd, a novel DiT-based framework that achieves identity-consistent and motion-faithful video generation in multi-character scenarios, enabling versatile applications such as arbitrary character animation and identity-motion recasting; 
(2) we introduce IILR and TSDA (integrated with AGF) to effectively eliminate identity entanglement and ensure high-fidelity identity-pose binding with superior generalization; 
(3) we curate the MCD-7K dataset and establish MCD-300, the first benchmark to feature dense scenarios with up to 9 characters, where AnyCrowd sets a new state-of-the-art for high-density, consistent MCA generation.

\section{Related Work}
\label{sec:related}

% \subsection{Diffusion Transformers for Video Generation}
\noindent\textbf{Diffusion Transformers for Video Generation.}
Diffusion models~\cite{ho2020denoising,song2020denoising} have become the dominant paradigm for video synthesis due to their powerful generative capabilities. In recent years, the architectural backbone of these models has transitioned from convolutional U-Net frameworks\cite{ho2022video,singer2022make,ho2022imagen,blattmann2023stable,guo2023animatediff} toward Diffusion Transformers (DiTs)~\cite{hong2022cogvideo,yang2024cogvideox,kong2024hunyuanvideo,wu2025hunyuanvideo,wan2025wan}, which offer superior scalability for processing large-scale datasets and enable the generation of longer, more complex video content.
A notable model is Wan~\cite{wan2025wan}, a versatile open-source DiT framework that excels in text/image-to-video tasks and adapts seamlessly to downstream applications. Leveraging Wan's generative priors, VACE~\cite{jiang2025vace} enables versatile controllable video generation tasks, including character animation. 
% Our work, AnyCrowd, builds upon VACE to achieve high-quality multi-character animation.
Building on this foundation, we introduce AnyCrowd for high-quality multi-character animation.

% \subsection{Controllable Character Animation}
\noindent\textbf{Single-Character Animation.} 
Controllable character animation has evolved from early GAN-based frameworks~\cite{dong2018soft,siarohin2019first,li2019dense,wu2021image,zhang2022exploring,zhao2022thin} to Diffusion-based paradigms \cite{hu2024animate,xu2024magicanimate,zhu2024champ,tan2024animate,hu2025animate,wang2025unianimate,jiang2025vace,tu2025stableanimator,cheng2025wan,shao2025interspatial,ding2025mtvcrafter,zhang2024mimicmotion,zhou2025realisdance,zhang2025flexiact}. Pioneering U-Net based Diffusion frameworks like Animate Anyone~\cite{hu2024animate} and MagicAnimate~\cite{xu2024magicanimate} employ ReferenceNet for identity injection and temporal attention for temporal consistency. Subsequent works like MimicMotion~\cite{zhang2024mimicmotion} and StableAnimator~\cite{tu2025stableanimator} focus on refining fine-grained details (e.g., faces and hands), while Animate-X~\cite{tan2024animate} optimizes motion representations for anthropomorphic character animation.
More recently, frameworks such as UniAnimate-DiT~\cite{wang2025unianimate}, RealisDance-DiT~\cite{zhou2025realisdance}, VACE~\cite{jiang2025vace}, and Wan-Animate~\cite{wan2025wan} adopt DiT backbones~\cite{peebles2023scalable}, leveraging their superior scalability to significantly enhance identity preservation and motion coherence. However, these methods remain primarily tailored for single-character animation and struggle to generalize to multi-character scenarios (MCA).

\noindent\textbf{Multi-Character Animation.}
Existing MCA methods primarily distinguish identities through specialized input encodings. For instance, Xue et al.~\cite{xuetowards}, Wang et al.~\cite{wang2025multi}, and DanceTogether~\cite{chen2025dancetogether} utilize depth-based order maps, learnable ID embeddings, or mask-pose adapters to associate identities with their corresponding poses. However, these approaches struggle to generalize to character counts beyond their training distribution. Furthermore, Wang et al.~\cite{wang2025multi} encounters difficulties in cross-driven animation—where reference images and pose sequences originate from different sources—limiting its practical applicability.
More recently, CoDance~\cite{tan2026codance} has achieved support for arbitrary character counts but remains focused on single-pose animation, struggling with the complex overlaps and interactions of dense scenes. While SCAIL~\cite{yan2025scail} addresses multi-character interactions, it remains prone to identity-pose mis-binding due to the absence of an explicit binding mechanism. In contrast, AnyCrowd incorporates Tri-Stage Decoupled Attention (TSDA) to explicitly bind identities to their respective motion tracks, enabling identity-consistent animation for an arbitrary number of characters, even in highly interactive scenarios.

% \subsection{Multi-Character Animation Generation}

\section{Methodology}
\label{sec:method}

\begin{figure}[tb]
  \centering
  \includegraphics[width=\textwidth]{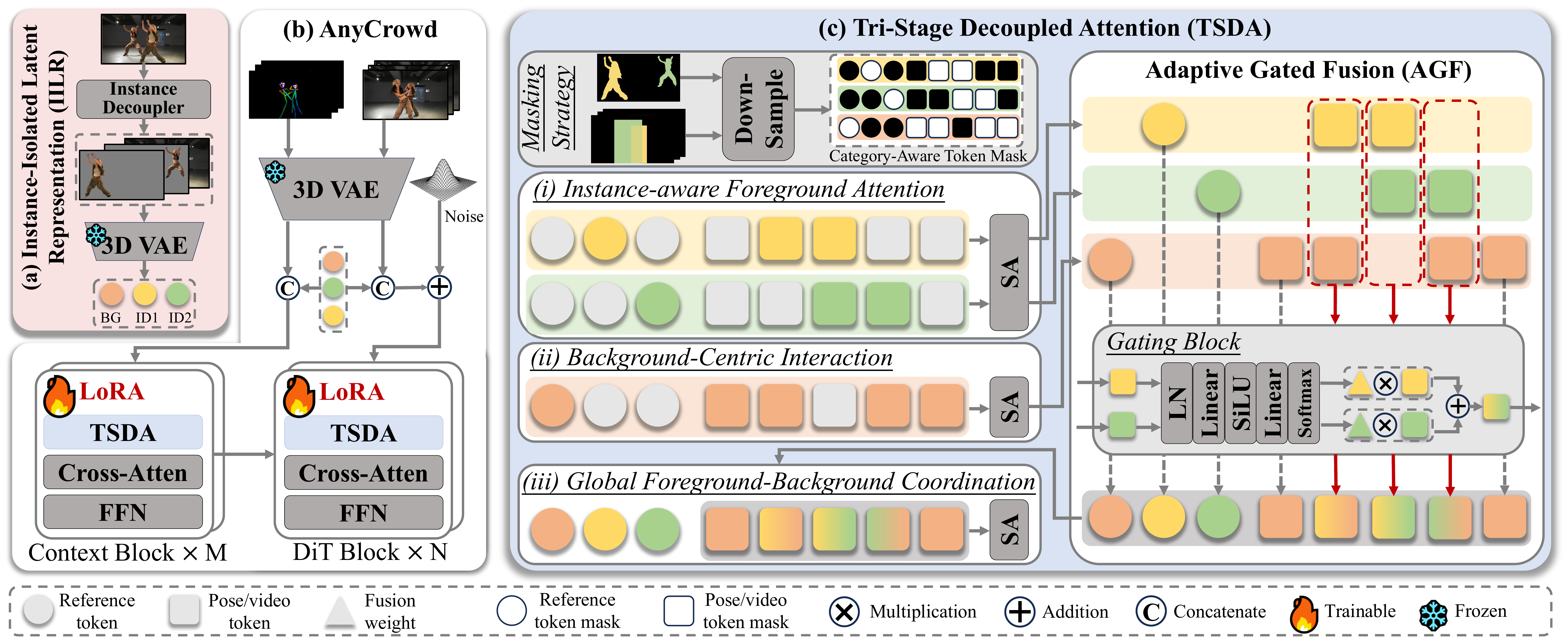}
  \caption{Overview of AnyCrowd. (a) Instance-Isolated Latent Representation (IILR): The reference image with $C$ identities is decoupled into $C+1$ isolated images and encoded into identity-decoupled reference tokens. (b) Architecture: AnyCrowd is built upon a dual-stream DiT architecture, where the Context and DiT branches process conditioning signals and perform iterative denoising. (c) Tri-Stage Decoupled Attention (TSDA): This mechanism facilitates explicit identity-pose binding during the self-attention process, incorporating an Adaptive Gated Fusion (AGF) module to adaptively fuse overlapping tokens from different categories.}
  \label{fig:framework}
\end{figure}

Given a reference image $I_{\mathrm{ref}}$ containing $C$ identities and a driving video $V_{\mathrm{drv}}$ depicting $C$ characters, Multi-Character Animation (MCA) aims to synthesize a video $V$ that preserves distinct identities while following the target motions. To this end, we propose \textbf{AnyCrowd}, a DiT-based framework (Sec.~\ref{subsec:anycrowd}) that introduces Instance-Isolated Latent Representation (IILR, Sec.~\ref{subsec:iilr}) for identity-decoupled encoding and Tri-Stage Decoupled Attention (TSDA, Sec.~\ref{subsec:tsda}) for explicit identity-pose binding. We further detail the training objectives in Sec.~\ref{subsec:objectives}.
% and introduce the newly curated dataset and benchmark in Sec.~\ref{subsec:data}. 
Fig.~\ref{fig:framework} illustrates the structural overview; while only two characters are depicted for clarity, AnyCrowd naturally scales to an arbitrary number of subjects.

\subsection{Framework of AnyCrowd}
\label{subsec:anycrowd}

Inspired by recent advancements in controllable video generation~\cite{jiang2025vace}, AnyCrowd is built upon a dual-stream Diffusion Transformer (DiT) architecture. It consists of a Context Branch to process conditioning signals (i.e., the reference image $I_{\mathrm{ref}}$ and the driving pose sequence $P$ extracted from $V_{\mathrm{drv}}$) and a DiT Branch to perform the iterative denoising of video tokens.
Specifically, AnyCrowd employs a pre-trained 3D VAE encoder to map $I_{\mathrm{ref}}$ and $P$ into reference tokens $\mathcal{Z}_{\mathrm{ref}}$ and pose tokens $\mathcal{Z}_\mathrm{p}$, respectively. 
During training, the ground-truth video $V_{\mathrm{drv}}$ is also encoded into video tokens $\mathcal{Z}_\mathrm{vid}$ using the same encoder. 
To provide appearance and structural guidance, the reference and pose tokens are concatenated along the temporal dimension and fed into the Context Branch. Simultaneously, the reference and video tokens are concatenated, perturbed with Gaussian noise, and sent to the DiT Branch. In this arrangement, the reference tokens serve as a temporal prefix to maintain appearance consistency throughout the generation.
The Context Branch consists of $M$ context blocks, while the DiT Branch consists of $N$ DiT blocks ($M < N$). In this asymmetric design, the output of each context block is injected into its corresponding DiT block to provide identity information and motion guidance for the denoising process. 
To adapt this architecture for multi-character animation, we incorporate two core modifications. First, AnyCrowd introduces \textbf{Instance-Isolated Latent Representation (IILR)} to replace standard reference encoding, providing identity-decoupled reference tokens. Furthermore, AnyCrowd reformulates the original self-attention in both the DiT and context blocks with a novel \textbf{Tri-Stage Decoupled Attention (TSDA)} to facilitate explicit identity-pose binding.
During the training phase, the 3D VAE remains frozen. We apply Low-Rank Adaptations~\cite{hu2022lora} (LoRAs) to the self-attention (SA), cross-attention, and feed-forward networks (FFN) in both branches to preserve the pre-trained generative priors. Notably, within TSDA, AnyCrowd further introduces an \textbf{Adaptive Gated Fusion (AGF)} with gating blocks, which are trained from scratch to learn the adaptive fusion strategies required for complex multi-character scenarios.
For the sake of concise notation, the term $\mathcal{Z}_{\mathrm{vid}}$ is hereafter used to collectively denote both the video tokens in the DiT branch and the pose tokens in the Context Branch.

\subsection{Instance-Isolated Latent Representation}
\label{subsec:iilr}

While holistic VAE encoding is effective for single-character scenarios, it exhibits inherent limitations in multi-character animation. The downsampling nature of VAE encoders causes the receptive fields of adjacent individuals to overlap in the latent space, resulting in severe identity entanglement. This issue is further exacerbated as the character density increases, where reduced spatial intervals between individuals lead to significant feature interference among reference tokens.
To resolve this, AnyCrowd introduces Instance-Isolated Latent Representation (IILR). As illustrated in Fig.~\ref{fig:framework}(a), the reference image $I_{\mathrm{ref}}$ is first processed by an Instance Decoupler to extract $C$ distinct characters and the background based on instance masks. These components are then projected onto $C+1$ separate images, each initialized with a neutral gray background and maintaining the same dimensions as $I_{\mathrm{ref}}$.
% Crucially, while the spatial layout of identities in $I_{\mathrm{ref}}$ does not inherently dictate their respective motion paths in $V_{\mathrm{drv}}$, we observe that a significant portion of our training samples exhibit coincidentally similar global positions across both domains. Such spatial consistency 
Crucially, although the spatial layout of identities in $I_{\mathrm{ref}}$ is theoretically independent of the motion paths in $V_{\mathrm{drv}}$, human-centric videos often exhibit a natural statistical bias where characters maintain consistent relative spatial orders. Such spatial correlations allow the network to potentially exploit coordinates as a shortcut for identity-pose mapping, rather than learning robust visual representations. To mitigate this, we perform a stochastic spatial reassignment of these isolated instances (excluding the background) within the Instance Decoupler. By rearranging characters into non-overlapping random positions, we disrupt the original coordinate dependency, forcing the model to establish associations based on intrinsic visual identity rather than inherited spatial locations.
These $C+1$ images are subsequently processed by the frozen 3D VAE encoder to produce $C$ identity reference tokens $\{\mathcal{Z}_{\mathrm{ref}}^1, \dots, \mathcal{Z}_{\mathrm{ref}}^C\}$ and one background reference token $\mathcal{Z}_{\mathrm{ref}}^{\mathrm{bg}}$. This mechanism ensures that each identity token is encoded in isolation and remains mutually invisible during the encoding stage, providing a disentangled foundation for the subsequent identity-pose binding in TSDA.

\subsection{Tri-Stage Decoupled Attention}
\label{subsec:tsda}

\noindent\textbf{Tri-Stage Attention Mechanism and Masking Strategy.}
Existing DiT-based methods~\cite{jiang2025vace,wang2025unianimate,cheng2025wan} typically employ global self-attention to learn correspondences among the reference tokens and video tokens. However, this approach is fundamentally limited by the inherent spatial stochasticity between reference character identities and driving pose tracks in multi-character scenarios. As discussed in Sec.~\ref{subsec:iilr}, the spatial ordering of identities in $I_{\mathrm{ref}}$ lacks a deterministic alignment with their corresponding trajectories in the driving frames; consequently, the global attention mechanism struggles to establish accurate matches without explicit identity-pose correspondence guidance. To address these challenges, AnyCrowd introduces the Tri-Stage Decoupled Attention (TSDA) mechanism, which decomposes the standard self-attention into three specialized phases to provide the necessary structural guidance for robust identity-pose binding.

Specifically, AnyCrowd first performs instance-aware foreground attention. For each character $i$, a dedicated reference token mask $\mathcal{M}_{\mathrm{ref}}^i$ and a video token mask $\mathcal{M}_{\mathrm{vid}}^i$ are utilized to isolate its respective tokens from its corresponding reference tokens $\mathcal{Z}_{\mathrm{ref}}^i$ and video tokens $\mathcal{Z}_{\mathrm{vid}}$. The self-attention calculation is restricted to the subset of tokens belonging to the same identity:
\begin{equation}
    \widehat{\mathcal{Z}}^{i}=\operatorname{SA}\left(\mathrm{Q}\left(\mathcal{Z}^{i}\right), \mathrm{K}\left(\mathcal{Z}^{i}\right), \mathrm{V}\left(\mathcal{Z}^{i}\right)\right), \quad  \mathcal{Z}^{i} = \{ z_j \in \mathcal{Z'}^i \mid m_{j}^i = 1, m_{j}^i \in \mathcal{M}^i\},
\end{equation}
where $\operatorname{SA}(\cdot,\cdot,\cdot)$ refers to the self-attention calculation, $\mathcal{Z'}^i = [\mathcal{Z}_{\mathrm{ref}}^i, \mathcal{Z}_{\mathrm{vid}}]$ denotes the concatenated token sequence, and  $\mathcal{M}^i = [\mathcal{M}^i_{\mathrm{ref}}, \mathcal{M}^i_{\mathrm{vid}}]$ represents the joint identity token mask.
By restricting the attention calculation to these isolated regions, AnyCrowd enforces a deterministic identity-to-pose mapping for each individual, thereby ensuring that each appearance is faithfully "implanted" into its respective motion track without interference from neighboring identities.
Subsequently, the framework conducts background-centric interaction to facilitate communication between the reference background tokens and the video background regions, which can be formulated as:
\begin{equation}
    \widehat{\mathcal{Z}}^{\mathrm{bg}}=\operatorname{SA}\left(\mathrm{Q}\left(\mathcal{Z}^{\mathrm{bg}}\right), \mathrm{K}\left(\mathcal{Z}^{\mathrm{bg}}\right), \mathrm{V}\left(\mathcal{Z}^{\mathrm{bg}}\right)\right), \quad \mathcal{Z}^{\mathrm{bg}} = \{ z_j \in \mathcal{Z'} \mid m_{j} = 1, m_{j} \in \mathcal{M}^{\mathrm{bg}}\},
\end{equation}
where $\mathcal{Z'} = [\mathcal{Z}_{\mathrm{ref}}^{\mathrm{bg}}, \mathcal{Z}_{\mathrm{vid}}]$ and $\mathcal{M}^{\mathrm{bg}} = [\mathcal{M}^{\mathrm{bg}}_{\mathrm{ref}}, \mathcal{M}^{\mathrm{bg}}_{\mathrm{vid}}]$.
This stage specifically focuses on preserving the environmental context and maintaining visual consistency between the static reference source and the synthesized dynamic frames.
Finally, AnyCrowd executes global foreground-background coordination. Once the identity and background information have been integrated into the video tokens, a final global self-attention is performed exclusively across all video tokens, which can be formulated as:
\begin{equation}
    \overline{\mathcal{Z}}_{\mathrm{vid}}=\operatorname{SA}\left(\mathrm{Q}\left(\widetilde{\mathcal{Z}}_{\mathrm{vid}}\right), \mathrm{K}\left(\widetilde{\mathcal{Z}}_{\mathrm{vid}}\right), \mathrm{V}\left(\widetilde{\mathcal{Z}}_{\mathrm{vid}}\right)\right),
\end{equation}
where $\widetilde{\mathcal{Z}}_{\mathrm{vid}}$ represents the identity-aware video tokens aggregated from the previous two stages. Crucially, the reference tokens are excluded in this phase. Re-introducing all reference tokens into a global interaction would allow features from different identities to cross-contaminate the video tokens once again, thereby disrupting the precise identity-to-pose mappings established in the preceding stages. This global interaction allows the now "identity-aware" video tokens to resolve complex scene dynamics—such as inter-character occlusions and foreground-background interactions—while maintaining the structural integrity of each individual's appearance.

The implementation of TSDA relies on specific token masks for both reference and video tokens, though their acquisition strategies differ to ensure model generalization during inference. As illustrated in Fig.~\ref{fig:framework}(c), for the reference tokens, AnyCrowd employs precise instance masks derived from $I_{\mathrm{ref}}$, which are downsampled to the latent space and reshaped as token masks. These masks define the categorical membership of each token, indicating which tokens correspond to a specific character identity or the background. In contrast, for the video tokens, AnyCrowd opts for Bounding Box (BBox) masks rather than fine-grained tracking masks from the driving video to prevent information leakage. Using precise masks during training would allow the model to learn a shortcut by merely copying the driving character's silhouette. This leads to a silhouette mismatch during inference when the driver’s morphology (e.g., hairstyles or clothing styles) differs significantly from the reference identity. By providing only coarse spatial guidance via BBoxes, the network is prevented from leveraging driving silhouettes as a shortcut, ensuring that the generated character's morphology is governed solely by the reference identity. 
% However, this BBox-based approach inevitably introduces overlapping regions where tokens are assigned to multiple categories—a challenge addressed by the Adaptive Gated Fusion (AGF) module described below.

\noindent\textbf{Adaptive Gated Fusion.}
While the first two stages of TSDA successfully establish precise identity-pose binding, the underlying BBox-based masking strategy inevitably introduces spatial overlaps within the video token domain. As illustrated by the red dashed boxes in Fig. 3(c), a video token position $j$ may result in multiple category-specific candidate tokens $\{\widehat{z}^{i}_{j}\}$ (e.g., from overlapping identities or foreground-background intersections). Naively aggregating these conflicting candidates triggers feature collision and identity ambiguity, where the synthesized appearance becomes a blurred mixture of multiple subjects.

To resolve this, AnyCrowd introduces the Adaptive Gated Fusion (AGF) module before the last stage of TSDA to perform a learnable arbitration among these candidates. Specifically, for a position $j$ that falls within the overlapping set of categories $\mathcal{C}_j$, the corresponding candidate tokens are fed into a shared Gating Block $\Phi(\cdot)$ to predict importance scores. Subsequently, a softmax operation is applied across these scores to obtain normalized fusion weights $w_{j}^{i}$, which are then used to aggregate the candidate tokens as follows:
\begin{equation}
\label{eq:fused}
    \widetilde{z}_{j}^{\mathrm{fused}} = \sum_{i \in \mathcal{C}_j} w_{j}^{i} \widehat{z}_{j}^{i}; \quad w_{j}^{i} = \frac{\exp(s_{j}^{i})}{\sum_{k \in \mathcal{C}_j} \exp(s_{j}^{k})}; \quad s_j^{i} = \Phi(\widehat{z}_{j}^{i}), i \in \mathcal{C}_j.
\end{equation}
Finally, these fused tokens, along with those from non-overlapping regions that bypass the AGF module, are aggregated to form the identity-aware video tokens $\widetilde{\mathcal{Z}}_{\mathrm{vid}}$, which are then fed into the final stage of TSDA for global coordination.

\subsection{Training Objectives}
\label{subsec:objectives}

To facilitate the learning of the Gating Block within TSDA, we introduce a supervision loss based on soft labels. Specifically, for each video token position $j$, a ground-truth label $y^i_j$ is derived by average pooling the high-resolution binary tracking instance masks of the driving video to the latent resolution. The resulting $y^i_j \in [0, 1]$ reflects the spatial occupancy ratio of category $i$ within the pixel patch corresponding to that latent position. 
A cross-entropy loss is then employed to align the predicted fusion weights $w^i_j$ (defined in Eq.~\ref{eq:fused}) with these soft labels:
\begin{equation}\mathcal{L}_{\mathrm{agf}} = - \frac{1}{|\Omega|} \sum_{j \in \Omega} \sum_{i \in \mathcal{C}_j} y^i_j \log(w^i_j),
\end{equation}
where $\Omega$ represents the set of video token indices in overlapping regions and $\mathcal{C}_j$ denotes the set of candidate categories at position $j$. By minimizing this objective, the Gating Block learns to accurately identify identity attribution from latent features, effectively resolving identity conflicts during inference even when precise masks are unavailable.

Beyond identity arbitration, AnyCrowd adopts the video generation loss from Wan2.1~\cite{wan2025wan}, which is grounded in Rectified Flows (RFs)~\cite{esser2024scaling}. This objective, denoted as $\mathcal{L}_{\mathrm{gen}}$, minimizes the mean squared error (MSE) between the predicted velocity field and the ground-truth velocity field defined by the flow matching path. The overall training objective for AnyCrowd is thus formulated as:
\begin{equation}
    \mathcal{L}_{\mathrm{total}} = \mathcal{L}_{\mathrm{gen}} + \lambda_{\mathrm{agf}} \mathcal{L}_{\mathrm{agf}},
\end{equation}
where $\lambda_{\mathrm{agf}}$ is a hyperparameter to balance the two loss terms and is set to 1.

\section{Experiments}
\label{sec:exp}

\begin{figure}[tb]
  \centering
  \includegraphics[width=\textwidth]{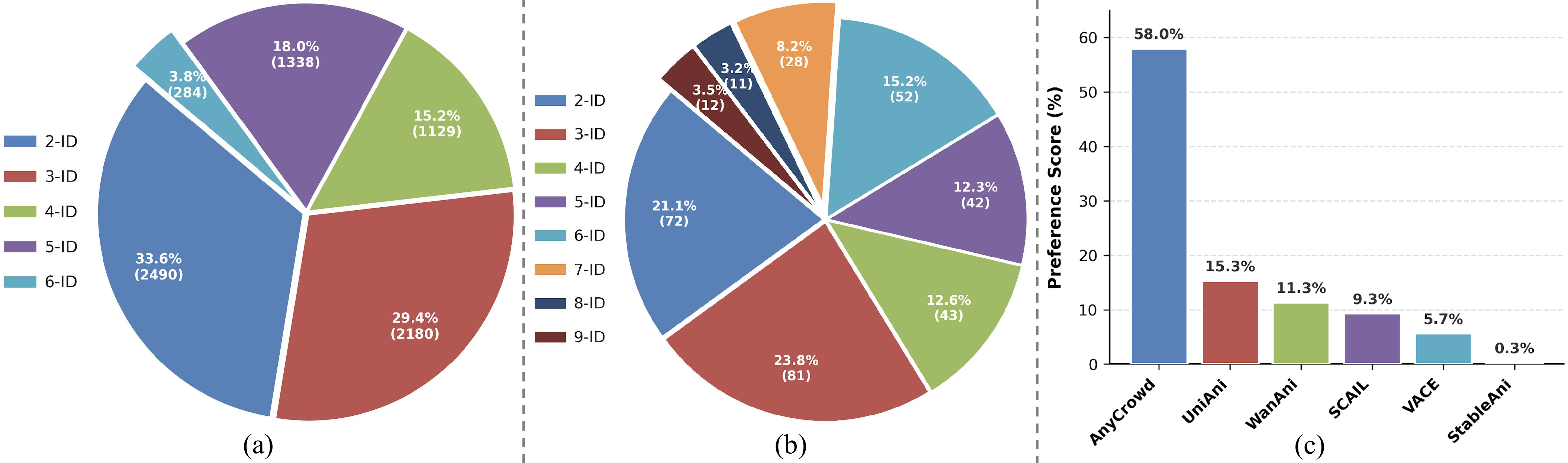}
  \caption{(a, b) Character number distribution of MCD-7K (train) and MCD-300 (test). (c) Preference results on MCD-300 under cross-driven setting.}
  \label{fig:dataset_he}
\end{figure}

% \subsection{Experiment Setup}
\noindent\textbf{Dataset and Benchmark.} To support multi-character training, we curate the Multi-Character-Dancing-7K (MCD-7K) dataset, containing 7,384 video clips (totaling $\sim$31.8 hours). Unlike existing datasets~\cite{jafarian2021learning,chen2025dancetogether,xuetowards,yan2025scail}, MCD-7K specifically features 2 to 6 performers with frequent spatial transitions and intricate inter-character interactions. Furthermore, to address the lack of large-scale evaluation protocols, we establish MCD-300, the first benchmark dedicated to high-density scenarios. It comprises 341 videos featuring 2 to 9 dancers, intentionally exceeding the character counts in our training set to evaluate out-of-distribution generalization (see Fig.~\ref{fig:dataset_he} (a,b) for dataset and benchmark distributions). 
% Detailed curation pipelines and dataset statistics are provided in the Appendix.

% \noindent\textbf{Implementation details.}
% We adopt Wan2.1-VACE-14B~\cite{jiang2025vace} as the backbone for AnyCrowd. To maintain the robust generative priors of the base model, we freeze all original backbone parameters and insert LoRA~\cite{hu2022lora} layers with a rank of $r=32$ into both the Context Blocks and DiT Blocks. These LoRA layers, together with the Gating Block in TSDA, serve as the only trainable components. During training, we randomly sample 49 consecutive frames from each video in the MCD-7K dataset (at roughly 25--30 FPS) and randomly select one additional frame as the reference image. All data are resized and center-cropped to a resolution of $480 \times 832$. AnyCrowd is optimized using the AdamW~\cite{loshchilov2017decoupled} optimizer with a learning rate of $1 \times 10^{-4}$ and a weight decay of $1 \times 10^{-2}$. The training is conducted on 7 NVIDIA H100 (80GB) GPUs with a batch size of 1 per GPU. We further employ BF16 mixed-precision training and gradient checkpointing offload to reduce the memory footprint, enabling stable training and convergence over a total of 20,000 iterations.

\noindent\textbf{Baselines.}
We compare AnyCrowd with several state-of-the-art open-source methods, including Wan-Animate~\cite{cheng2025wan}, VACE~\cite{jiang2025vace}, Uni-Animate~\cite{wang2025unianimate}, and StableAnimator~\cite{tu2025stableanimator}, all retrained on our MCD-7K dataset for a fair comparison. We also evaluate against the multi-character baseline SCAIL~\cite{yan2025scail}. Since its training code is not released, we use their released 14B model directly for comparison.

\noindent\textbf{Evaluation Protocol.}
Following~\cite{yan2025scail}, we evaluate AnyCrowd in two modes: \emph{self-driven} and \emph{cross-driven}. In the self-driven setting, the reference image and driving pose sequence are extracted from the same video. We conduct evaluations on all 341 clips from MCD-300, employing frame-wise metrics (PSNR~\cite{hore2010image}, SSIM~\cite{wang2004image}, LPIPS~\cite{zhang2018unreasonable}) and distribution-wise metrics (FID~\cite{heusel2017gans}, FID-VID~\cite{balaji2019conditional}, FVD~\cite{unterthiner2018towards}) to measure reconstruction quality. Furthermore, to assess instance-level identity consistency, we calculate the Cosine Similarity of CLIP~\cite{radford2021learning} and DINOv2~\cite{oquab2023dinov2} features between the reference character and the corresponding generated character across 10 uniformly sampled frames from the generated videos. We utilize ground-truth instance tracking masks to precisely extract character regions from both reference images and generated frames.
In the cross-driven setting, where the reference and driving sources are extracted from different videos with distinct identities, we partition MCD-300 by character count and randomly pair different sources to form a test set of 272 samples, which better reflect real-world multi-character animation and test model generalization. Given the absence of ground truth in this mode, we rely on FID, FID-VID, and FVD to assess the distribution similarity between generated and real videos, alongside a user study to evaluate visual realism and identity consistency.\footnote{Comprehensive details regarding the dataset curation, implementation specifics, and user study protocols are provided in the Appendix.}

\begin{figure}[tb]
  \centering
  \includegraphics[width=\textwidth]{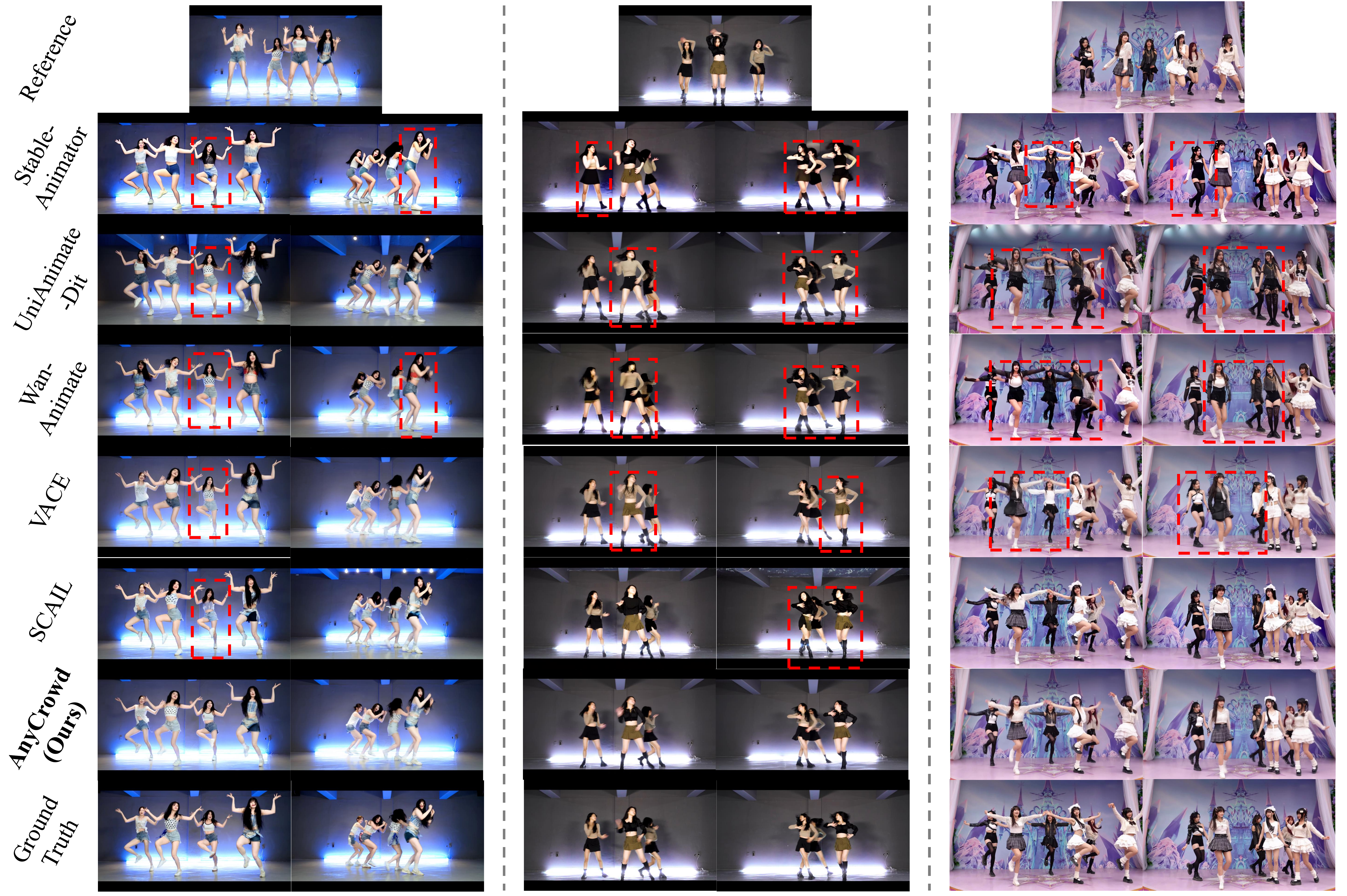}
  \caption{Qualitative comparison on MCD-300 under the self-driven setting. Red dashed boxes highlight typical artifacts in baseline results. Please zoom in for details.
  }
  \label{fig:selfdriven_results}
\end{figure}

\begin{figure}[tb]
  \centering
  \includegraphics[width=\textwidth]{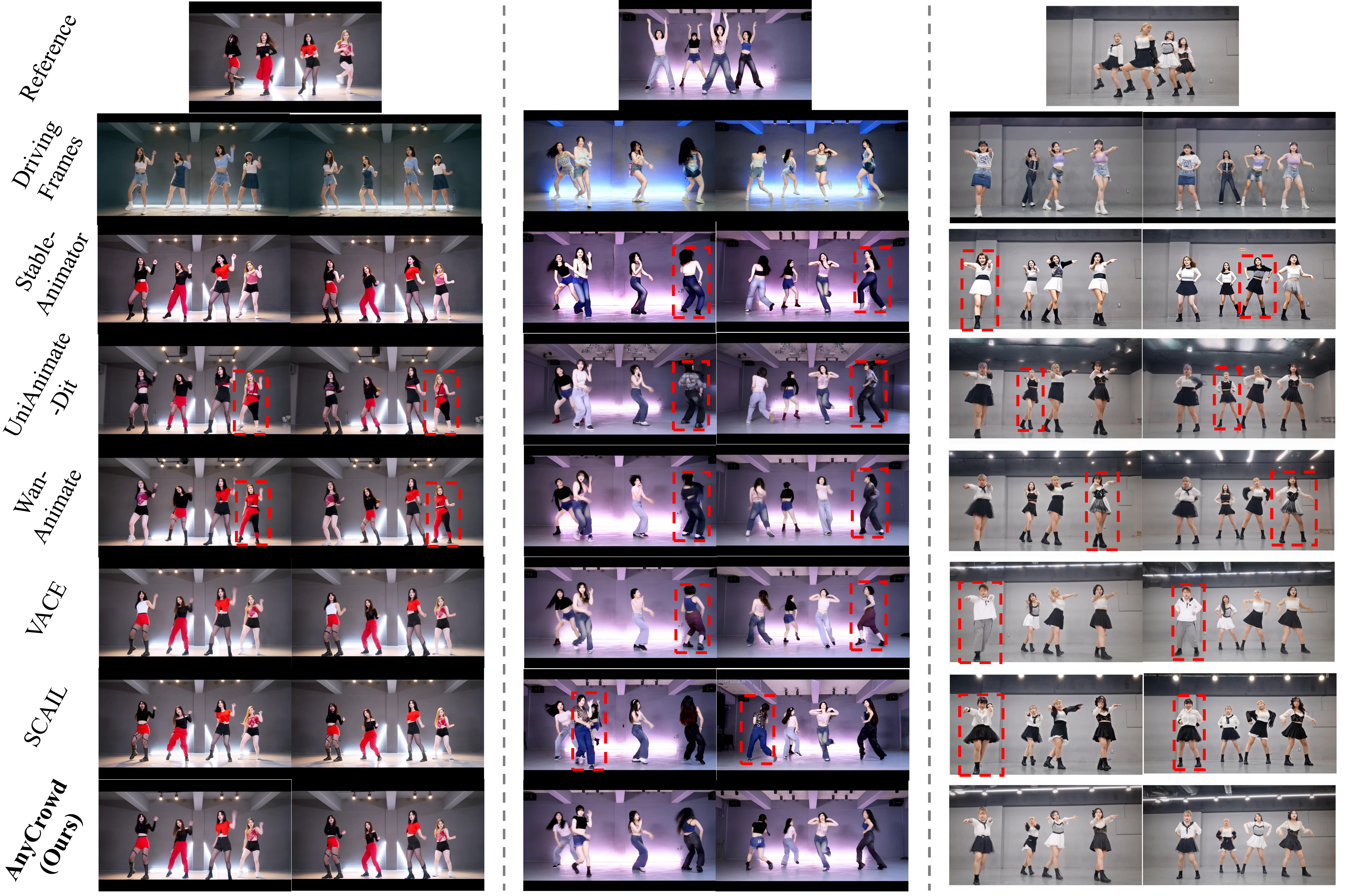}
  \caption{Qualitative comparison on MCD-300 under the cross-driven setting. Red dashed boxes highlight typical artifacts in baseline results. Please zoom in for details.
  }
  \label{fig:crossdriven_results}
\end{figure}

\begin{figure}[tb]
  \centering
  \includegraphics[width=\textwidth]{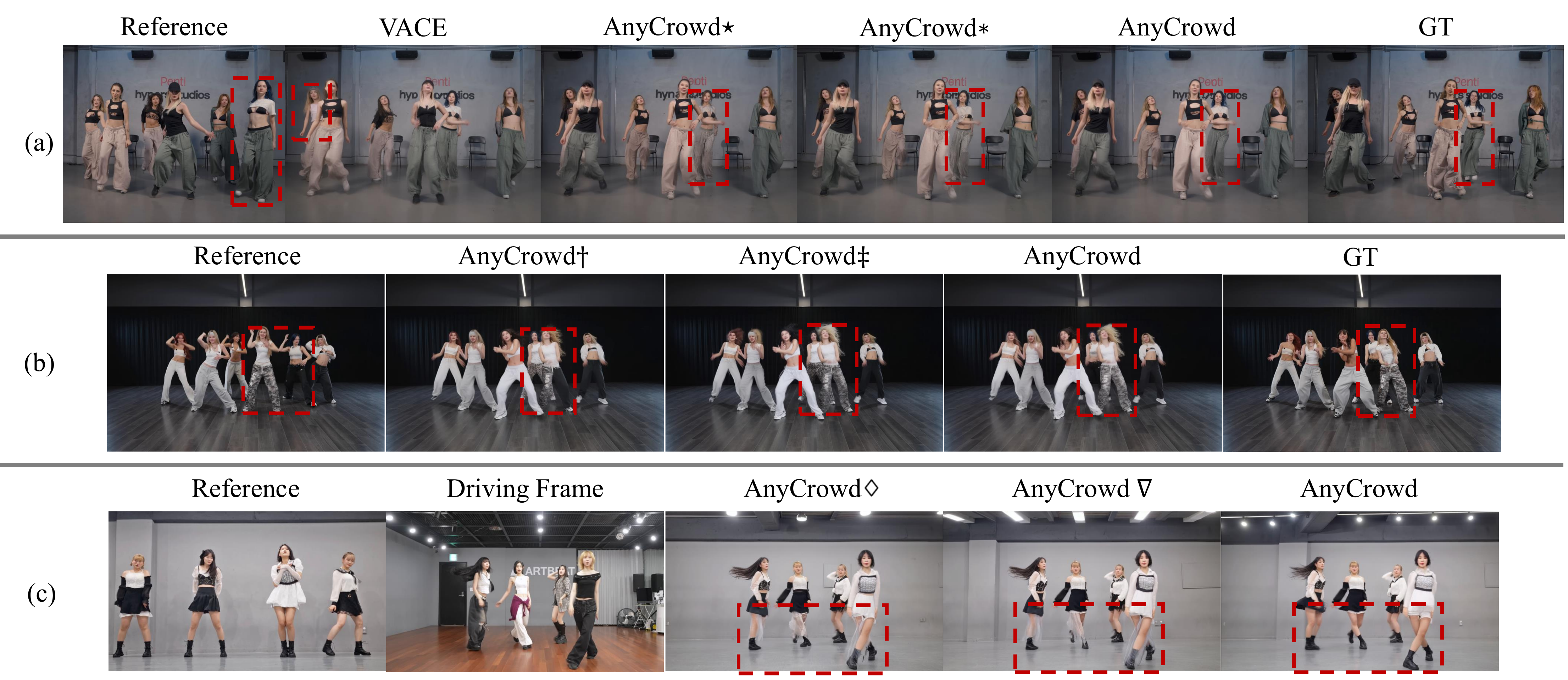}
  \caption{Ablation study. (a) Effectiveness of TSDA and AGF. (b) Necessity of IILR. (c) Influence of masking strategies (under cross-driven setting). Please zoom in for details.
  }
  \label{fig:abl}
\end{figure}

\begin{table}[t]
\centering
\tiny
% \scriptsize
\setlength{\tabcolsep}{1.5pt}
\renewcommand{\arraystretch}{1.3}
\caption{Quantitative comparison of AnyCrowd with baselines on MCD-300.}                   
\label{tab:quan_comp}
        
\begin{tabular}{lcccccccc|ccc}
\toprule
\multirow{3}{*}{\textbf{Methods}} &
% \multirow[c]{3}{*}{\parbox[c]{2cm}{\centering\textbf{Methods}}} &
\multicolumn{8}{c|}{\textbf{Self-Driven}} &
\multicolumn{3}{c}{\textbf{Cross-Driven}} \\
\cmidrule(lr){2-9}\cmidrule(lr){10-12}

% ===== 新增：指标类别分组行 =====
& \multicolumn{3}{c}{\textbf{Frame-wise}} 
& \multicolumn{3}{c}{\textbf{Distribution-wise}} 
& \multicolumn{2}{c|}{\textbf{ID Consistency}}
& \multicolumn{3}{c}{\textbf{Distribution-wise}} \\
\cmidrule(lr){2-4}\cmidrule(lr){5-7}\cmidrule(lr){8-9}\cmidrule(lr){10-12}

% ===== 原来的指标名行 =====
& PSNR$\uparrow$ & SSIM$\uparrow$ & LPIPS$\downarrow$
& FID$\downarrow$ & FID-VID$\downarrow$ & FVD$\downarrow$
& CLIP-I$\uparrow$ & DINO-I$\uparrow$
& FID$\downarrow$ & FID-VID$\downarrow$ & FVD$\downarrow$ \\
\midrule
StableAnimator~\cite{tu2025stableanimator} & 16.84 & 0.659 & 0.302 & 26.29 & 29.87 & 321.8 & 0.756 & 0.581 & 42.13 & 45.89 & 596.3 \\
UniAnimate-DiT~\cite{wang2025unianimate} & 16.12 & 0.641 & 0.340 & 16.86 & 21.90 & 153.6 & 0.797 & 0.637 & 32.46 & 40.04 & 354.0 \\
Wan-Animate~\cite{cheng2025wan}      & 18.16 & 0.688 & 0.272 & 17.03 & 18.76 & 160.5 & 0.770 & 0.568 & 33.21 & 39.32 & 431.4 \\
VACE~\cite{jiang2025vace}                 & 17.67 & 0.683 & 0.276 & 14.03 & 17.36 & 115.6 & 0.793 & 0.630 & 29.94 & 37.69 & 316.9 \\
SCAIL-14B~\cite{yan2025scail}      & 14.91 & 0.586 & 0.371 & 15.82 & 20.43 & 186.7 & 0.784 & 0.605 & 34.11 & 41.61 & 444.7 \\
\midrule
\textbf{AnyCrowd (Ours)}        & \textbf{18.96} & \textbf{0.698} & \textbf{0.249} & \textbf{12.16} & \textbf{14.35} & \textbf{96.45} & \textbf{0.814} & \textbf{0.683} & \textbf{28.83} & \textbf{36.34} & \textbf{306.3} \\
\bottomrule
\end{tabular}
\end{table}

\subsection{Experiment Results}
\noindent\textbf{Quantitative results.}
As reported in Tab.~\ref{tab:quan_comp}, our method consistently outperforms all baseline models across all metrics under the self-driven setting. These results demonstrate that AnyCrowd achieves superior video generation quality. In particular, the significant lead in ID consistency metrics validates that our pose-identity binding mechanism effectively enhances identity stability in multi-character scenarios. Furthermore, in the more challenging cross-driven setting, both the superior distribution-wise metrics and the human preference results (as shown in Fig.~\ref{fig:dataset_he} (c)) collectively underscore the effectiveness and robustness of AnyCrowd in real-world scenarios. Overall, these results attest to the superiority of AnyCrowd for multi-character animation.

\noindent\textbf{Qualitative results.}
Fig.~\ref{fig:selfdriven_results} and Fig.~\ref{fig:crossdriven_results} illustrate the qualitative comparisons under self-driven and cross-driven settings, respectively. As shown in the self-driven results, existing methods (e.g., UniAnimate~\cite{wang2025unianimate}, VACE~\cite{jiang2025vace}, and Wan-Animate~\cite{cheng2025wan}) frequently exhibit ID-pose misbinding, where the spatial correspondence between characters and poses deviates from the ground truth (e.g., the first two cases in Fig.~\ref{fig:selfdriven_results}). Furthermore, baselines suffer from either distorted identity appearance (e.g., SCAIL~\cite{yan2025scail}, Wan-Animate, and StableAnimator~\cite{tu2025stableanimator} in the 1st case) or identity bleeding (e.g., VACE in the 2nd case), where visual features of different characters are incorrectly blended.
These challenges persist in the cross-driven setting (Fig.~\ref{fig:crossdriven_results}), where baselines fail to maintain identity fidelity. In contrast, AnyCrowd consistently produces superior results with accurate ID-pose binding and distinct character identities. This robustness is primarily attributed to our IILR module, which facilitates identity-isolated feature encoding, as well as the TSDA and AGF modules that ensure precise ID-to-pose mapping. Consequently, AnyCrowd ensures that each character is synthesized with high fidelity while remaining unaffected by neighboring identities.

\begin{table}[t]
\centering
\tiny
% \scriptsize
\setlength{\tabcolsep}{1.5pt}
\renewcommand{\arraystretch}{1.3}

\caption{Ablation study of core compoments on MCD-300.}          
\label{tab:ablation}

\begin{tabular}{c l cccccccc|ccc} % ← 新增了一列 group：最左侧 c
\toprule
\multirow{3}{*}{\textbf{}} & % ← 新增表头
\multirow{3}{*}{\textbf{Methods}} &
\multicolumn{8}{c|}{\textbf{Self-Driven}} &
\multicolumn{3}{c}{\textbf{Cross-Driven}} \\
\cmidrule(lr){3-10}\cmidrule(lr){11-13}

& & \multicolumn{3}{c}{\textbf{Frame-wise}} 
& \multicolumn{3}{c}{\textbf{Distribution-wise}} 
& \multicolumn{2}{c|}{\textbf{ID Consistency}}
& \multicolumn{3}{c}{\textbf{Distribution-wise}} \\
\cmidrule(lr){3-5}\cmidrule(lr){6-8}\cmidrule(lr){9-10}\cmidrule(lr){11-13}

& & PSNR$\uparrow$ & SSIM$\uparrow$ & LPIPS$\downarrow$
& FID$\downarrow$ & FID-VID$\downarrow$ & FVD$\downarrow$
& CLIP-I$\uparrow$ & DINO-I$\uparrow$
& FID$\downarrow$ & FID-VID$\downarrow$ & FVD$\downarrow$ \\
\midrule

% ------- (a) 覆盖 3 行 -------
\multirow{3}{*}{(a)} & Baseline~\cite{jiang2025vace} & 17.67 & 0.683 & 0.276 & 14.03 & 17.36 & 115.6 & 0.793 & 0.630 & 29.94 & 37.69 & 316.9 \\
                    & AnyCrowd$\star$                & 18.76 & 0.692 & 0.257 & 12.73 & 14.85 & 99.22 & 0.811 & 0.678 & 29.42 & 36.80 & 307.4 \\
                    & AnyCrowd$\ast$                 & 18.88 & 0.693 & 0.251 & 12.57 & 14.66 & 98.26 & 0.811 & 0.681 & 29.11 & 37.15 & 309.1 \\
\midrule

% ------- (b) 覆盖 2 行 -------
\multirow{2}{*}{(b)} & AnyCrowd$\dagger$            & 18.47 & 0.686 & 0.258 & 12.62 & 14.58 & 97.48 & 0.813 & 0.682 & 29.13 & 36.98 & 310.9 \\
                    & AnyCrowd$\ddagger$           & 18.67 & 0.692 &  0.259 & 12.85 & 14.81 & 104.4 & 0.809 & 0.676 & 30.43 & 36.89 & 309.8 \\
\midrule

% ------- (c) 覆盖 2 行 -------
\multirow{2}{*}{(c)} & AnyCrowd$\Diamond$           & 18.60 & 0.679 & 0.268 & 11.91 & 14.99 & 97.00 & 0.812 & 0.679 & 28.95 & 37.07 & 310.2 \\
                    & AnyCrowd$\nabla$             & \textbf{19.12} & 0.695 & 0.250 & \textbf{11.64} & 14.46 & 97.36 & 0.813 & 0.681 & 29.26 & 37.24 & 308.7 \\
\midrule

% ------- Full（可留空或写 Full）-------
{}                  & \textbf{AnyCrowd}            & 18.96 & \textbf{0.698} & \textbf{0.249} & 12.16 & \textbf{14.35} & \textbf{96.45} & \textbf{0.814} & \textbf{0.683} & \textbf{28.83} & \textbf{36.34} & \textbf{306.3} \\
\bottomrule
\end{tabular}

\end{table}

\subsection{Ablation Studies}
% We conduct a series of ablation experiments (Tab. \ref{tab:ablation} and Fig.~\ref{fig:abl}) to evaluate: (a) the effectiveness of Tri-Stage Decoupled Attention (TSDA) and Adaptive Gated Fusion (AGF) for robust identity-pose binding in multi-character animation; (b) the capability of Instance-Isolated Latent Representation (IILR) in preventing identity entanglement; and (c) the necessity of using bounding box masks for video token extraction within TSDA to enhance generalization in cross-driven scenarios.
\noindent\textbf{Effectiveness of TSDA and AGF.}
We evaluate three variants: (i) Baseline (VACE~\cite{jiang2025vace}), which employs standard global self-attention without identity-pose binding; (ii) AnyCrowd${\star}$, utilizing TSDA but resolving overlapping tokens via max-selection; and (iii) AnyCrowd${\ast}$, which replaces our adaptive gated fusion (AGF) with mean-pooling. As shown in Tab.~\ref{tab:ablation} (a), all AnyCrowd-based models consistently outperform the baseline, validating the fundamental role of identity-pose binding within the TSDA module. Moreover, the performance degradation in AnyCrowd${\star}$ and AnyCrowd${\ast}$ underscores the necessity of AGF. Qualitatively, while the baseline fails to maintain basic identity-pose alignment and suffers from distorted identity appearance, the variants without AGF struggle with identity entanglement in high-density regions (Fig.~\ref{fig:abl} (a), red dashed boxes). In contrast, the complete AnyCrowd model achieves precise ID-pose binding and preserves accurate identity characteristics even in crowded scenarios.

\noindent\textbf{Necessity of IILR.}
We evaluate two variants: (i) AnyCrowd${\dagger}$, which employs globale reference encoding by feeding the entire image into the VAE encoder, leading to identity-entangled latent tokens; and (ii) AnyCrowd${\ddagger}$, which performs per-instance encoding but omits the stochastic spatial reassignment, leaving each character's tokens spatially tied to its original coordinates. The performance drop observed in Tab.~\ref{tab:ablation} (b) illustrates the necessity of IILR. Qualitatively, as shown in Fig.~\ref{fig:abl} (b), both variants suffer from identity bleeding with neighboring characters. 

% In contrast, the full AnyCrowd effectively prevents identity entanglement and preserves distinct identity characteristics.
% These variants are compared against the full AnyCrowd, which leverages instance-isolated encoding combined with random shuffling.

\noindent\textbf{Analysis of Masking Strategies.}
We evaluate two additional configurations: (i) AnyCrowd${\Diamond}$, which utilizes precise ground-truth (GT) tracking masks from the driving video; and (ii) AnyCrowd${\nabla}$, which employs dilated GT masks for coarser boundaries. As shown in Tab.~\ref{tab:ablation} (c), these variants show marginal gains in certain metrics under the self-driven setting, this is due to information leakage from target silhouettes. However, such leakage is detrimental in cross-driven scenarios, leading to performance degradation. As illustrated in Fig.~\ref{fig:abl} (c), these variants over-rely on driving geometries, causing shape-related artifacts (e.g., generating pants-like texture for characters wearing skirts). In contrast, our bounding box strategy effectively avoids these shape priors, ensuring faithful identity and outfit preservation during cross-driven transfer.

\section{Conclusion}
\label{sec:conclu}
We present AnyCrowd, a novel DiT-based framework for animating an arbitrary number of characters. By introducing the IILR and TSDA (integrated with AGF) modules, our method eliminates identity entanglement and ensures precise ID-pose binding in crowded scenarios. We further introduce the MCD-7K dataset and the MCD-300 benchmark to facilitate research in high-density character generation. Extensive experiments demonstrate that AnyCrowd sets a new state-of-the-art, offering superior generalization and fidelity for versatile applications. Limitations and broader impacts are discussed in the Appendix.
%we will discuss the limitation and social impact in the Appendix.

% \clearpage\mbox{}Page \thepage\ of the manuscript.
% \clearpage\mbox{}Page \thepage\ of the manuscript.
% \clearpage\mbox{}Page \thepage\ of the manuscript.
% \clearpage\mbox{}Page \thepage\ of the manuscript.
% \clearpage\mbox{}Page \thepage\ of the manuscript. This is the last page.
% Now we have reached the maximum length of an ECCV \ECCVyear{} submission (excluding references and acknowledgements).
% References should start immediately after the main text, but can continue past p.\ 14 if needed. 
\clearpage  % TODO FINAL: This \clearpage needs to be removed from both review and camera-ready versions.

% \section*{Acknowledgements}
% Please insert your acknowledgments here.

\bibliographystyle{splncs04}
\bibliography{main}

\clearpage
\appendix

\section{Appendix Introduction}
\label{sec:intro}
This supplementary document provides additional technical details, extended experimental results, and broader discussions to complement the main manuscript of AnyCrowd. We first provide a comprehensive description of the MCD-7K dataset and the MCD-300 benchmark in Sec.~\ref{sec:mcd}, detailing the automated curation pipeline and key dataset characteristics. This is followed by an elaboration on our evaluation protocol in Sec.~\ref{sec:evaluation}, which specifies the implementation of evaluation metrics and the setup of the user preference study. To ensure reproducibility, we present extensive implementation details in Sec.~\ref{sec:implementation}, covering the specific architectures of the newly introduced modules, hyperparameters, and training schedules. 
Furthermore, we provide a wealth of additional qualitative comparisons and versatile applications in Sec.~\ref{sec:results}. Finally, we conclude with a discussion on the limitations and broader impact of our work on generative AI research and the creative industry in Sec.~\ref{sec:limitation}.

\section{MCD-7K Dataset and MCD-300 Benchmark}
\label{sec:mcd}

\subsection{MCD-7K Dataset}

To facilitate training, we curate the Multi-Character-Dancing-7K (\textbf{MCD-7K}) dataset, which consists of video clips with resolutions ranging from 1K to 4K collected from online sources. This dataset is meticulously designed to encompass a rich spectrum of multi-person dynamics, featuring 2 to 6 performers per clip engaged in a wide variety of dance styles, such as K-Pop, folk dances, aerobics, zumba, ballroom dancing, and figure skating. By capturing such a diverse range of motion types, MCD-7K provides a comprehensive testbed for modeling frequent spatial transitions and intricate inter-character interactions in complex multi-human scenarios.

The construction of the MCD-7K dataset is achieved through a multi-stage, semi-automated pipeline that transforms raw web-sourced data into high-quality, annotated video content. We initially curate a collection of 5K high-definition source videos, which are segmented into approximately 50K video clips (ranging from 3 to 20 seconds each) using PySceneDetect. To ensure data quality and structural consistency, the pipeline follows a comprehensive workflow:

\begin{enumerate}
    \item \textbf{Detection and Estimation:} For every frame, we employ YOLOx~\cite{ge2021yolox} to obtain precise bounding boxes (bboxes) for each character, alongside DWPose~\cite{yang2023effective} to extract high-fidelity 2D human poses.
    \item \textbf{Filtering and Selection:} To handle dynamic noise such as occlusions, off-screen movements, or incidental background characters, we define the character count for each clip as the median number of individuals detected across all frames. Clips containing fewer than 2 or more than 6 individuals are strictly discarded to maintain high-density requirements.
    \item \textbf{Anchor-based Mask Tracking:} To establish a reliable starting point for tracking, we first identify the subset of frames whose character count matches the clip's defined label. From these candidates, we select an 'anchor frame' characterized by the maximum aggregate bbox area and minimal inter-bbox overlap, ensuring optimal visibility for all performers. Starting from this frame, we utilize the SAMURAI~\cite{yang2024samurai} algorithm to track each character's mask both forward and backward, merging these trajectories to obtain temporally consistent tracking masks for the entire clip.
    \item \textbf{Identity-Pose Binding:} We perform cross-modal alignment between the SAMURAI-generated tracking masks and the YOLOx-derived bboxes. By calculating the Intersection over Union (IoU) ratio, each 2D pose is reliably associated with a unique identity track.
\end{enumerate}
Following this automated process, we conduct a rigorous manual inspection to filter out samples with inaccurate pose estimations or tracking failures. This curation effort results in a final dataset of 7,384 video clips, totaling approximately 31.78 hours of multi-person dancing content. Fig.~\ref{fig:dataset_appendix}(a) shows representative samples from the MCD-7K dataset.

\subsection{MCD-300 Benchmark}
Furthermore, as existing benchmarks~\cite{jafarian2021learning,chen2025dancetogether,xuetowards,yan2025scail} primarily focus on single or low-density (2--3 people) scenarios, we introduce \textbf{MCD-300}, a more rigorous benchmark designed for complex, high-occupancy environments. It comprises 341 high-definition dancing videos featuring between 2 and 9 performers. Notably, the character count in MCD-300 intentionally exceeds that of our training set (MCD-7K) to evaluate the \textit{out-of-distribution generalization} of the models. The data curation and annotation process for MCD-300 strictly follows the same multi-stage pipeline as MCD-7K, including automated detection, anchor-based tracking, and identity-pose binding, all supplemented by meticulous manual verification to ensure ground-truth accuracy. 

As summarized in Tab.~\ref{tab:dataset_comparison}, MCD-300 offers significantly higher character density compared to existing datasets, providing a more demanding testbed for evaluating identity consistency and motion fidelity. Visual samples of MCD-300 are illustrated in Fig.~\ref{fig:dataset_appendix}(b), showcasing its unique challenges such as frequent position swapping and dense inter-character interactions in crowded scenes.

% 记得在导言区添加 \usepackage{multirow}
\begin{table*}[t]
\centering
\setlength{\tabcolsep}{6pt} % 稍微调大了一点间距，因为列数变多了
\caption{Quantitative comparison of benchmarks in the character animation field. While existing datasets primarily focus on single or low-density scenarios, our MCD-300 introduces a large-scale, high-density testbed specifically designed to advance the state-of-the-art in multi-character animation and motion fidelity in crowded environments.}
\label{tab:dataset_comparison}
\begin{tabular}{llcc}
\toprule
Category & Benchmark & Persons / Clip & Total Clips \\ 
\midrule
\multirow{4}{*}{Single-character} 
    & DisPose~\cite{li2025dispose} & 1 & 30 \\
    & HumanVid~\cite{wang2024humanvid} & 1 & 80 \\
    & UBC-Fashion~\cite{DBLP:journals/corr/abs-1910-09139} & 1 & 100  \\
    & TikTokDataset~\cite{jafarian2021learning} & 1 & 340  \\
\midrule
\multirow{4}{*}{Multi-character} 
    & Multi-Character~\cite{xuetowards} & 2 & 20 \\
    & DanceTogEval-100~\cite{chen2025dancetogether} & 2 & 100 \\
    & Studio-Bench~\cite{yan2025scail} & 1--2 & 130 \\
    & \textbf{MCD-300 (Ours)} & \textbf{2--9} & \textbf{341} \\ 
\bottomrule
\end{tabular}
\end{table*}

\section{More Evaluation Protocol Details}
\label{sec:evaluation}

\subsection{Evaluation Metrics Details}

% \textbf{Settings and Applicability.} 
As introduced in the main text, we evaluate our model under two scenarios: \emph{self-driven} and \emph{cross-driven} generation. 
For the \emph{self-driven} setting, where ground-truth (GT) videos are available, we employ a comprehensive suite of metrics covering frame-wise, distribution-wise, and identity-consistency evaluation. 
In contrast, the \emph{cross-driven} setting lacks GT references; thus, we primarily rely on distribution-wise metrics to assess the visual quality and temporal coherence of the generated results.

\noindent\textbf{Frame-wise Metrics.}
These metrics evaluate pixel-level or perceptual alignment between the generated frames $\hat{I}$ and ground-truth (GT) frames $I$.

\begin{itemize}
    \item \textbf{PSNR $\uparrow$:} Peak Signal-to-Noise Ratio (PSNR)~\cite{hore2010image} measures the ratio between the maximum possible power of a signal and the power of corrupting noise that affects the fidelity of its representation. In the context of video generation, it evaluates the pixel-level similarity between $\hat{I}$ and $I$. It is defined based on the Mean Squared Error (MSE):
    \begin{equation}
       \mathrm{MSE} = \frac{1}{H \times W} \sum_{i=1}^{H} \sum_{j=1}^{W} [I(i,j) - \hat{I}(i,j)]^2,
    \end{equation}
    \begin{equation}
      \mathrm{PSNR} = 10 \cdot \log_{10} \left( \frac{255^2}{\mathrm{MSE}} \right),
    \end{equation}
    where $H$ and $W$ denote the height and width of the image, and 255 is the maximum pixel intensity for 8-bit images. A higher PSNR value indicates superior pixel-level reconstruction quality.

    \item \textbf{SSIM $\uparrow$}: Structural Similarity Index Measure (SSIM)~\cite{wang2004image} assesses the perceived similarity between two images by capturing the degradation of structural information. Unlike pixel-level errors, SSIM models the human visual system by considering three comparative measurements: luminance ($l$), contrast ($c$), and structure ($s$). The SSIM index is defined as:
    \begin{equation}
        \text{SSIM}(I, \hat{I}) = [l(I, \hat{I})]^\alpha \cdot [c(I, \hat{I})]^\beta \cdot [s(I, \hat{I})]^\gamma,
    \end{equation}
    where $\alpha, \beta, \gamma > 0$ are parameters that adjust the relative importance of these components (typically set to 1). For image patches $I$ and $\hat{I}$, it is commonly calculated as:
    \begin{equation}
        \text{SSIM}(I, \hat{I}) = \frac{(2\mu_I\mu_{\hat{I}} + c_1)(2\sigma_{I\hat{I}} + c_2)}{(\mu_I^2 + \mu_{\hat{I}}^2 + c_1)(\sigma_I^2 + \sigma_{\hat{I}}^2 + c_2)},
    \end{equation}
    where $\mu_I$ and $\mu_{\hat{I}}$ denote the mean values, $\sigma_I^2$ and $\sigma_{\hat{I}}^2$ are the variances, $\sigma_{I\hat{I}}$ is the covariance, and $c_1, c_2$ are small constants used to stabilize the division. An SSIM score ranges from 0 to 1, where 1 indicates an identical structural match to the ground-truth image.

    \item \textbf{LPIPS $\downarrow$}: Learned Perceptual Image Patch Similarity (LPIPS)~\cite{zhang2018unreasonable} assesses the perceptual similarity between two images by computing the distance in the deep feature space of a pre-trained network. Unlike pixel-based metrics, LPIPS leverages feature activations to capture human-perceptual patterns. Given an image $I$ and its generated counterpart $\hat{I}$, LPIPS is defined as:
    \begin{equation}
        \text{LPIPS}(I, \hat{I}) = \sum_{l} \frac{1}{H_l W_l} \sum_{i,j} \| w_l \odot (\phi_l(I)(i,j) - \phi_l(\hat{I})(i,j)) \|_2^2,
    \end{equation}
    where $\phi_l$ denotes the feature extraction function at the $l$-th layer of a pre-trained network (e.g., VGG-16~\cite{simonyan2014very}), $w_l$ represents the learnable scaling weights for the $l$-th layer, and $H_l, W_l$ are the spatial dimensions of the feature map. A lower LPIPS score indicates higher perceptual similarity to the GT image.

\end{itemize}

\noindent\textbf{Distribution-wise Metrics.}
These metrics quantify the statistical divergence between the feature distributions of generated frames/videos and ground-truth (GT) counterparts.

\begin{itemize}
    \item \textbf{FID $\downarrow$}: Fréchet Inception Distance (FID)~\cite{heusel2017gans} measures the discrepancy between the feature distributions of generated frames and GT frames. It approximates the Inception-v3 feature activations (typically the 2048-D pool3 features)~\cite{szegedy2016rethinking} of real and generated images as multivariate Gaussians. Given $\mathcal{N}(\mu_{r}, \Sigma_{r})$ for real frames and $\mathcal{N}(\mu_{g}, \Sigma_{g})$ for generated frames, FID is defined as:
    \begin{equation}
        \mathrm{FID} = \|\mu_{r} - \mu_{g}\|_2^2 + \mathrm{Tr}\!\left(\Sigma_{r} + \Sigma_{g} - 2\left(\Sigma_{r}\Sigma_{g}\right)^{\frac{1}{2}}\right),
    \end{equation}
    where $\mu$ and $\Sigma$ denote the mean vector and covariance matrix of the Inception features, $\mathrm{Tr}(\cdot)$ is the trace operator, and $(\cdot)^{1/2}$ denotes the matrix square root. Lower FID indicates that the generated distribution is closer to the real data distribution.

    \item \textbf{FID-VID $\downarrow$}: To evaluate distribution alignment at the clip level, we compute a video-level Fréchet Inception Distance (FID-VID)~\cite{balaji2019conditional} by first pooling frame-wise Inception-v3 features over time. Given a video clip of $T$ frames $\{\mathbf{I}_t\}_{t=1}^{T}$, the clip-level feature is
    \begin{equation}
        \mathbf{f}_{\text{video}}=\frac{1}{T}\sum_{t=1}^{T}\phi(\mathbf{I}_t),
    \end{equation}
    where $\phi(\cdot)$ extracts features using a pre-trained Inception-v3 network~\cite{szegedy2016rethinking}. We then fit multivariate Gaussians to the pooled features of real and generated videos, denoted by $\mathcal{N}(\mu_r,\Sigma_r)$ and $\mathcal{N}(\mu_g,\Sigma_g)$, respectively, and compute
    \begin{equation}
        \mathrm{FID\text{-}VID}=\|\mu_r-\mu_g\|_2^2+\mathrm{Tr}\!\left(\Sigma_r+\Sigma_g-2(\Sigma_r\Sigma_g)^{\frac{1}{2}}\right).
    \end{equation}
    Lower values indicate that the generated videos are closer to real videos in terms of their clip-level (time-averaged) visual feature distribution.

    \item \textbf{FVD $\downarrow$}: 
    Fréchet Video Distance (FVD)~\cite{unterthiner2018towards} measures the distributional distance between generated video clips and GT video clips in a spatio-temporal feature space. Unlike frame-based metrics, FVD uses a pre-trained 3D CNN (commonly an I3D network~\cite{Carreira_2017_CVPR}, e.g., pre-trained on Kinetics) to extract features that capture both appearance and temporal dynamics. Assuming the extracted features of real and generated videos follow multivariate Gaussians $\mathcal{N}(\mu_{r}, \Sigma_{r})$ and $\mathcal{N}(\mu_{g}, \Sigma_{g})$, FVD is defined as:
    \begin{equation}
        \mathrm{FVD}=\|\mu_{r}-\mu_{g}\|_2^2+\mathrm{Tr}\!\left(\Sigma_{r}+\Sigma_{g}-2\left(\Sigma_{r}\Sigma_{g}\right)^{\frac{1}{2}}\right).
    \end{equation}
    Lower FVD indicates that generated videos are closer to real videos in terms of spatio-temporal feature distributions.

\end{itemize}

\noindent\textbf{Identity Consistency Metrics.}
These metrics evaluate how well a method preserves each character’s visual identity (e.g., hairstyle, clothing) throughout a generated video, especially under pose, viewpoint, and illumination changes.

\begin{itemize}
    \item \textbf{CLIP-I $\uparrow$}: To measure instance-level identity consistency in multi-character videos, we compute the cosine similarity between CLIP image embeddings~\cite{radford2021learning} of each character's reference appearance and its tracked regions over time, and then average across characters. Let $I_{\mathrm{ref}}$ be the reference frame and $M_{\mathrm{ref}}^{(c)}$ be the reference mask for character $c\in\{1,\dots,C\}$. Let $\hat{I}_t$ and $\hat{M}_t^{(c)}$ denote the generated frame and the tracked mask at time $t$, respectively. We extract character crops via $\mathcal{R}(\cdot)$ (masked extraction followed by square bounding-box cropping with padding), and encode them using the CLIP visual encoder $\phi_{\mathrm{CLIP}}(\cdot)$ (e.g., ViT-L/14). Define the per-frame similarity for character $c$ at time $t$ as:
    \begin{equation}
        s_t^{(c)}=\frac{\phi_{\mathrm{CLIP}}\!\left(\mathcal{R}(I_{\mathrm{ref}}, M_{\mathrm{ref}}^{(c)})\right)^{\top}\phi_{\mathrm{CLIP}}\!\left(\mathcal{R}(\hat{I}_{t}, \hat{M}_{t}^{(c)})\right)}{\left\|\phi_{\mathrm{CLIP}}\!\left(\mathcal{R}(I_{\mathrm{ref}}, M_{\mathrm{ref}}^{(c)})\right)\right\|_{2}\, \left\|\phi_{\mathrm{CLIP}}\!\left(\mathcal{R}(\hat{I}_{t}, \hat{M}_{t}^{(c)})\right)\right\|_{2}}.
    \end{equation}
    For each character, we average over the set of sampled frames $\mathcal{T}$ where the tracked region is valid (non-empty), denoted by $\mathcal{T}_c \subseteq \mathcal{T}$:
    \begin{equation}
        \bar{s}^{(c)}=\frac{1}{|\mathcal{T}_c|}\sum_{t\in\mathcal{T}_c} s_t^{(c)}.
    \end{equation}
    Finally, CLIP-I is obtained by averaging over all characters:
    \begin{equation}
        \mathrm{CLIP\text{-}I}=\frac{1}{C}\sum_{c=1}^{C}\bar{s}^{(c)}.
    \end{equation}
    Higher CLIP-I indicates better preservation of character identity across the generated video.

    \item \textbf{DINO-I $\uparrow$}: We similarly define DINO-I by replacing the CLIP encoder with a self-supervised DINOv2 visual encoder $\phi_{\mathrm{DINO}}(\cdot)$~\cite{oquab2023dinov2} (e.g., ViT-L/14). Using the same character crops $\mathcal{R}(\cdot)$ and sampled frames $\mathcal{T}$, we compute the per-frame cosine similarity:
    \begin{equation}
    d_t^{(c)}=\frac{\phi_{\mathrm{DINO}}\!\left(\mathcal{R}(I_{\mathrm{ref}}, M_{\mathrm{ref}}^{(c)})\right)^{\top}\phi_{\mathrm{DINO}}\!\left(\mathcal{R}(\hat{I}_{t}, \hat{M}_{t}^{(c)})\right)}{\left\|\phi_{\mathrm{DINO}}\!\left(\mathcal{R}(I_{\mathrm{ref}}, M_{\mathrm{ref}}^{(c)})\right)\right\|_{2}\,\left\|\phi_{\mathrm{DINO}}\!\left(\mathcal{R}(\hat{I}_{t},\hat{M}_{t}^{(c)})\right)\right\|_{2}}.
    \end{equation}
    We average over valid frames for each character $c$:
    \begin{equation}
        \bar{d}^{(c)}=\frac{1}{|\mathcal{T}_c|}\sum_{t\in\mathcal{T}_c} d_t^{(c)},
    \end{equation}
    and then average across characters:
    \begin{equation}
        \mathrm{DINO\text{-}I}=\frac{1}{C}\sum_{c=1}^{C}\bar{d}^{(c)}.
    \end{equation}
    Higher DINO-I indicates stronger identity consistency of multiple characters throughout the generated video.

\end{itemize}

\subsection{User Study Setting}
\begin{figure}[tb]
  \centering
  \includegraphics[width=\textwidth]{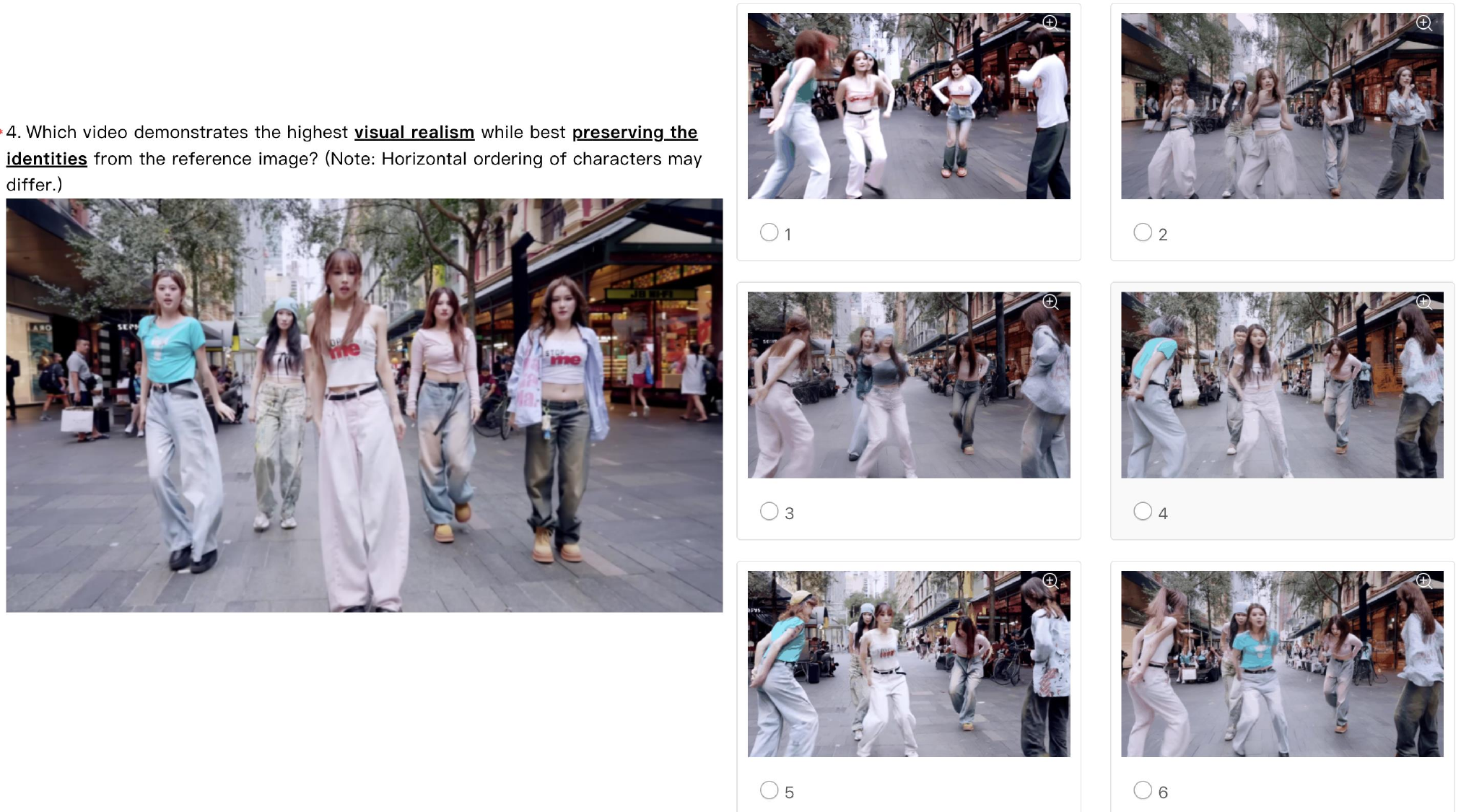}
  \caption{The Interface of User Study Questionnaire.}
  \label{fig:user_study}
\end{figure}

\begin{figure}[tb]
  \centering
  \includegraphics[width=\textwidth]{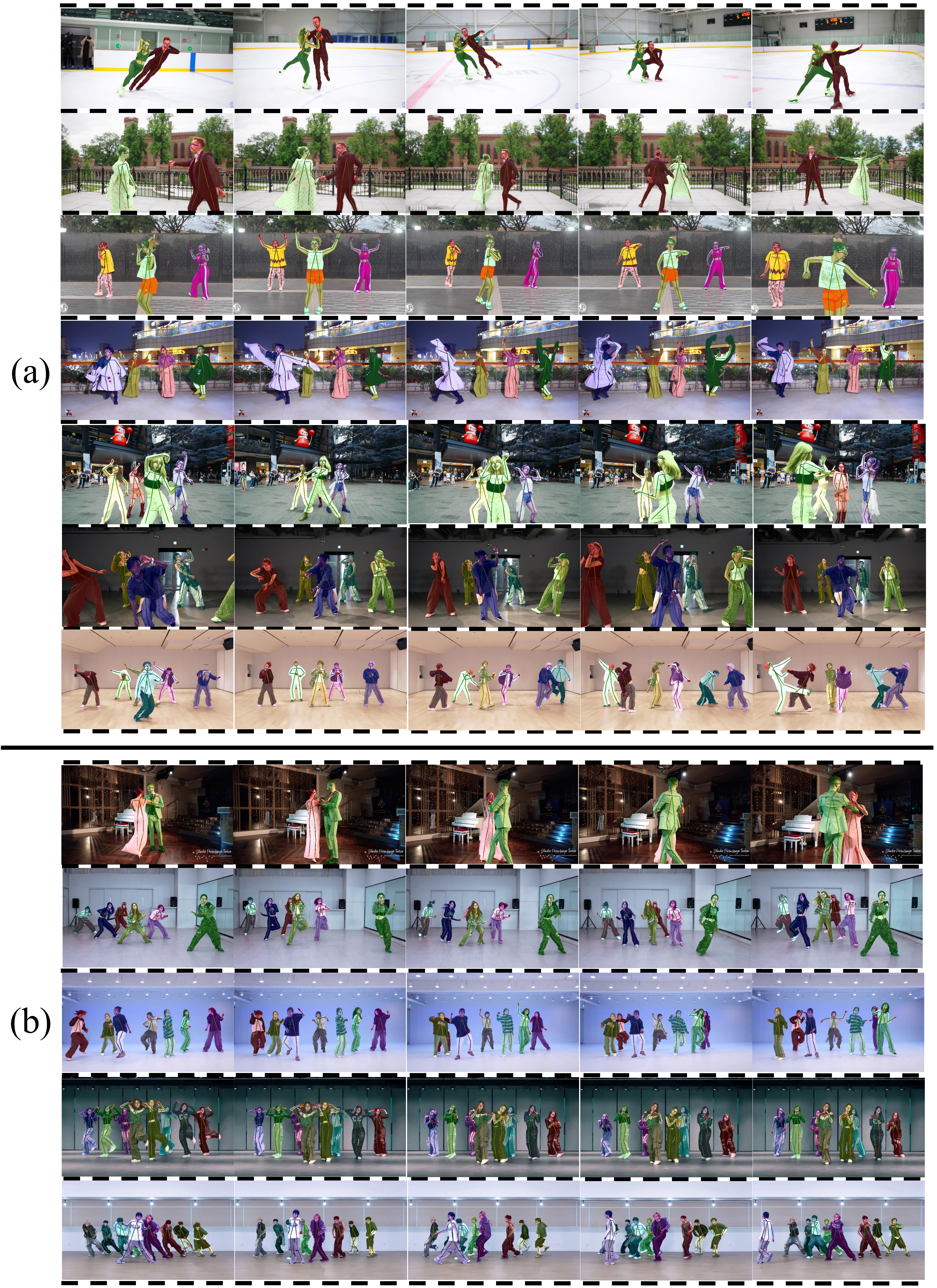}
  \caption{(a) showcases the diversity of the MCD-7K with 2--6 performers across various dance styles, while (b) highlights high-density scenarios with 6--9 performers from MCD-300 to evaluate model performance under out-of-distribution conditions. Tracking masks and 2D pose annotations are superimposed for visualization.}
  \label{fig:dataset_appendix}
\end{figure}

To further evaluate the perceptual quality of our method under the \emph{cross-driven} setting, we conduct a comprehensive user preference study. We randomly select 15 test samples from the 272 clips in the cross-driven setting. For each sample, we generate videos using AnyCrowd and five state-of-the-art baseline methods: StableAnimator~\cite{tu2025stableanimator}, UniAnimate-DiT~\cite{wang2025unianimate}, VACE~\cite{jiang2025vace}, Wan-Animate~\cite{cheng2025wan}, and Scail~\cite{yan2025scail}.
We recruit 20 volunteers to participate in the study, which consists of 15 questions. In each question, a reference image is provided alongside six generated videos, with their order randomized to eliminate selection bias. Participants are asked to select the best result based on two primary criteria: 1) \textbf{Visual Realism}, which assesses the naturalness of character motions and the absence of visual artifacts; and 2) \textbf{Identity Consistency}, which evaluates how well the generated character preserves the identity and appearance features of the person in the reference image. The interface of our user study questionnaire is illustrated in Fig.~\ref{fig:user_study}.

\section{Implementation  Details}
\label{sec:implementation}

\noindent\textbf{Architecture of Gating Block.}
To enable the adaptive fusion of multi-source information, we introduce a Gating Block that dynamically predicts fusion weights for each character identity. The block processes input features through a sequential architecture to determine the importance of different categories—specifically distinct identities and the background—for each token within overlapping regions. The architectural stack consists of a Layer Normalization (LN) layer, a linear projection, an SiLU activation, a second linear projection, and a final Softmax layer. For each token in the overlapping region, the Gating Block predicts a set of fusion weights, one for each category. These weights are applied to the corresponding input features via element-wise multiplication. Finally, the modulated features are aggregated through an element-wise summation to generate the refined output.

\noindent\textbf{Experimental Setup Details.}
We adopt Wan2.1-VACE-14B~\cite{jiang2025vace} as the backbone for AnyCrowd. To maintain the robust generative priors of the base model, we freeze all original backbone parameters and insert LoRA~\cite{hu2022lora} layers with a rank of $r=32$ into both the Context Blocks and DiT Blocks. These LoRA layers, together with the Gating Block in Tri-Stage Decoupled Attention (TSDA), serve as the only trainable components. During training, we randomly sample 49 consecutive frames from each video in the MCD-7K dataset (at roughly 25--30 FPS) and randomly select one additional frame as the reference image. All data are resized and center-cropped to a resolution of $480 \times 832$. AnyCrowd is optimized using the AdamW~\cite{loshchilov2017decoupled} optimizer with a learning rate of $1 \times 10^{-4}$ and a weight decay of $1 \times 10^{-2}$. 
The training is conducted on 7 NVIDIA H100 (80GB) GPUs with a batch size of 1 per GPU. The model is trained for a total of 18,000 iterations until convergence.

% We further employ BF16 mixed-precision training and gradient checkpointing offload to reduce the memory footprint, enabling stable training and convergence over a total of 18,000 iterations.

\noindent\textbf{Memory-Efficient Training Strategy.}
To reduce memory consumption during training, we adopt several complementary strategies. First, we pre-extract text embeddings using the T5 encoder~\cite{raffel2020exploring} offline and load them directly during optimization, eliminating the need to load the full T5 model into GPU memory. Since all video samples share the same text prompt, this further simplifies the data pipeline and reduces redundant computation. Second, we employ BF16 mixed-precision training to halve the memory footprint of model parameters and activations. Third, we enable gradient checkpointing to offload intermediate activations, trading off computation for memory savings. 

\noindent\textbf{Code and Models.} The training and inference code, along with the pre-trained weights, will be available.

\begin{figure}[tb]
  \centering
  \includegraphics[width=\textwidth]{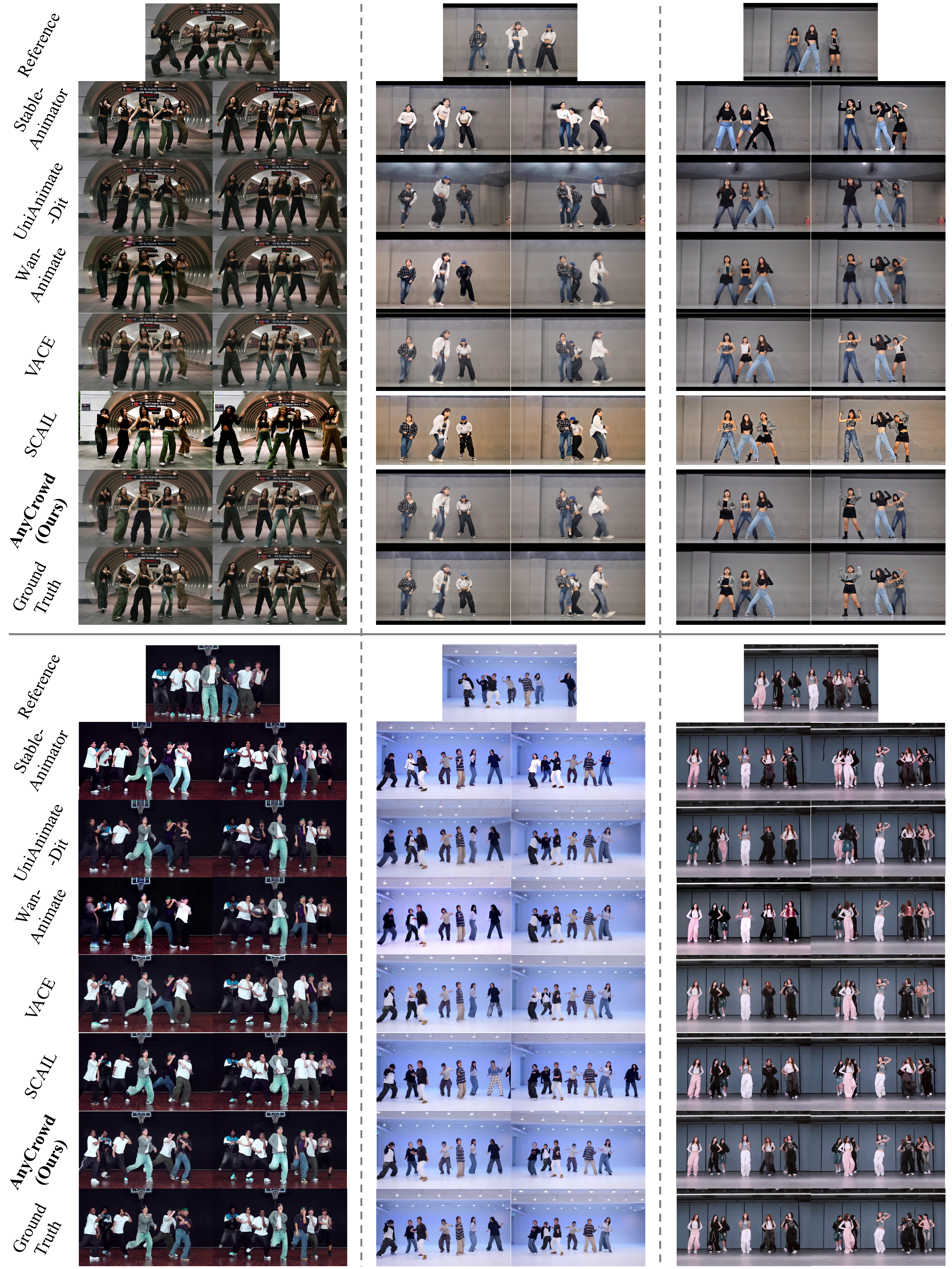}
  \caption{More qualitative comparisons on MCD-300 under the \emph{self-driven} setting. Please zoom in for details.}
  \label{fig:selfdriven_comp}
\end{figure}

\begin{figure}[tb]
  \centering
  \includegraphics[width=\textwidth]{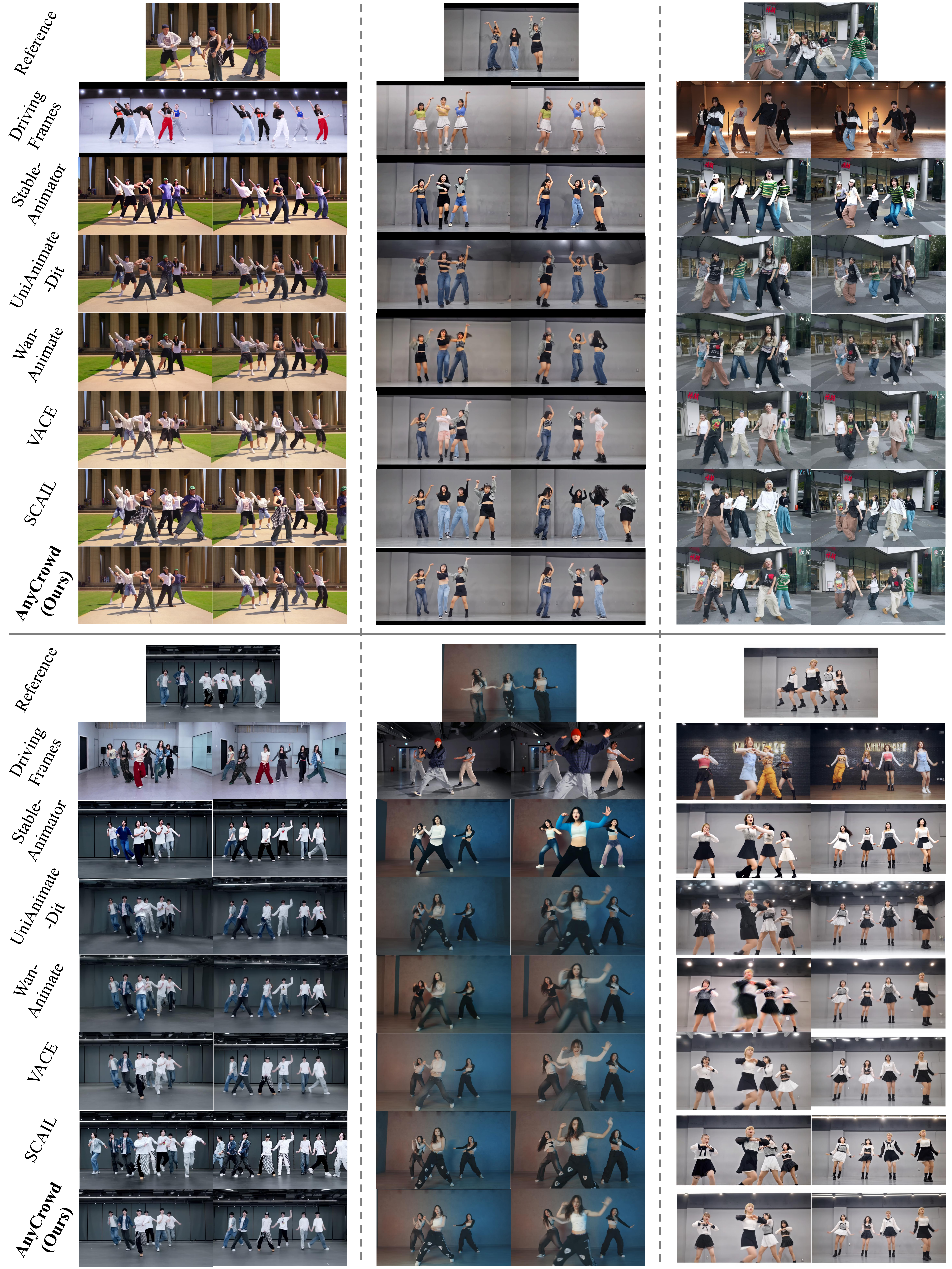}
  \caption{More qualitative comparisons on MCD-300 under the \emph{cross-driven} setting. Please zoom in for details.}
  \label{fig:crossdriven_comp}
\end{figure}

\begin{figure}[tb]
  \centering
  \includegraphics[width=\textwidth]{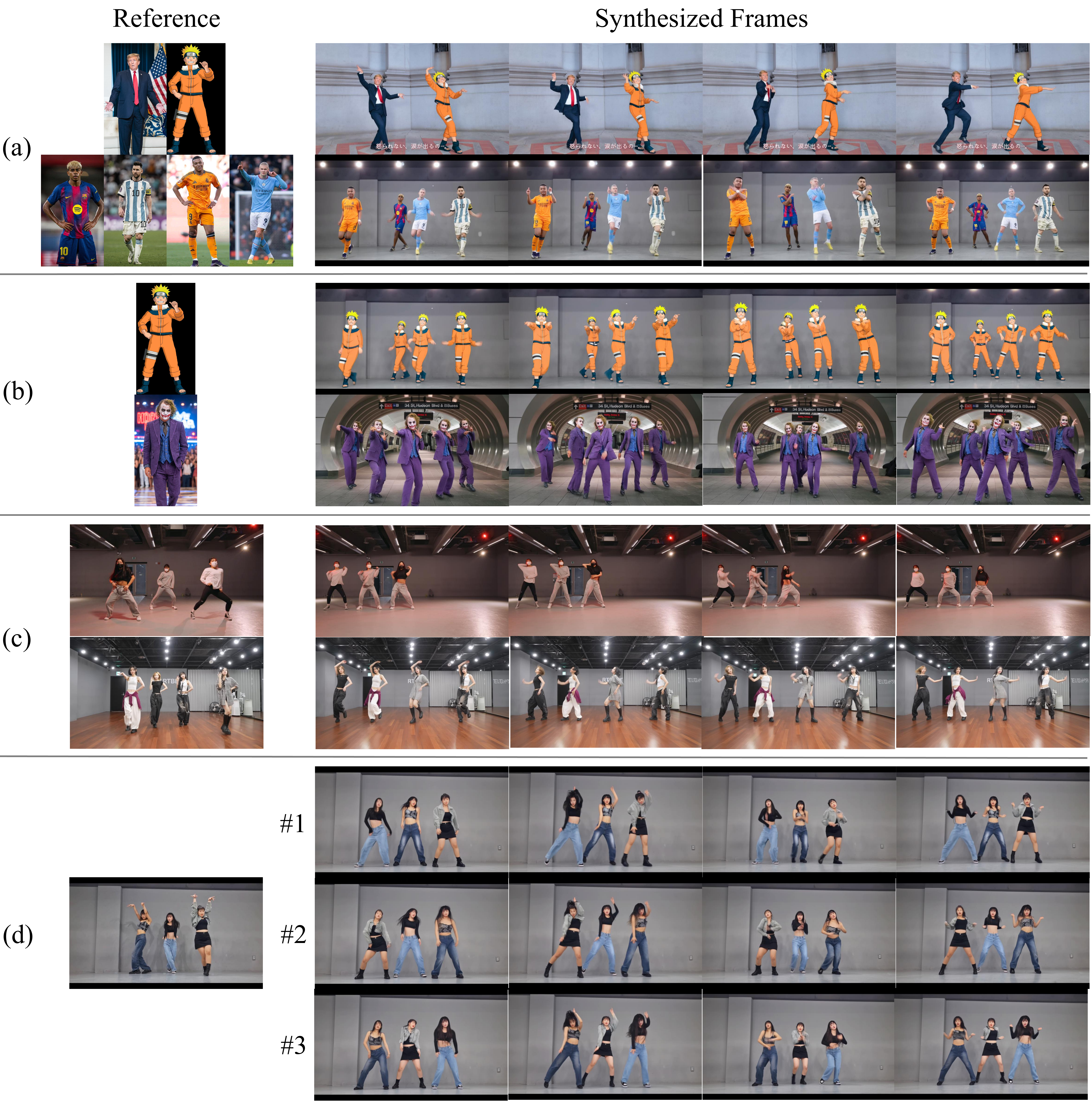}
  \caption{More application results of AnyCrowd. (a) Flexible Identity Input: AnyCrowd supports an arbitrary number of reference images from different sources. (b) One-to-Many Animation: A single reference character can be driven by diverse pose sequences. (c) Many-to-One Animation: Multiple distinct identities are animated simultaneously by a single driving pose sequence. (d) Arbitrary Pose-ID Mapping: AnyCrowd enables customized animation by reassigning specific identities to arbitrary target poses.}
  \label{fig:applications}
\end{figure}

\begin{figure}[tb]
  \centering
  \includegraphics[width=\textwidth]{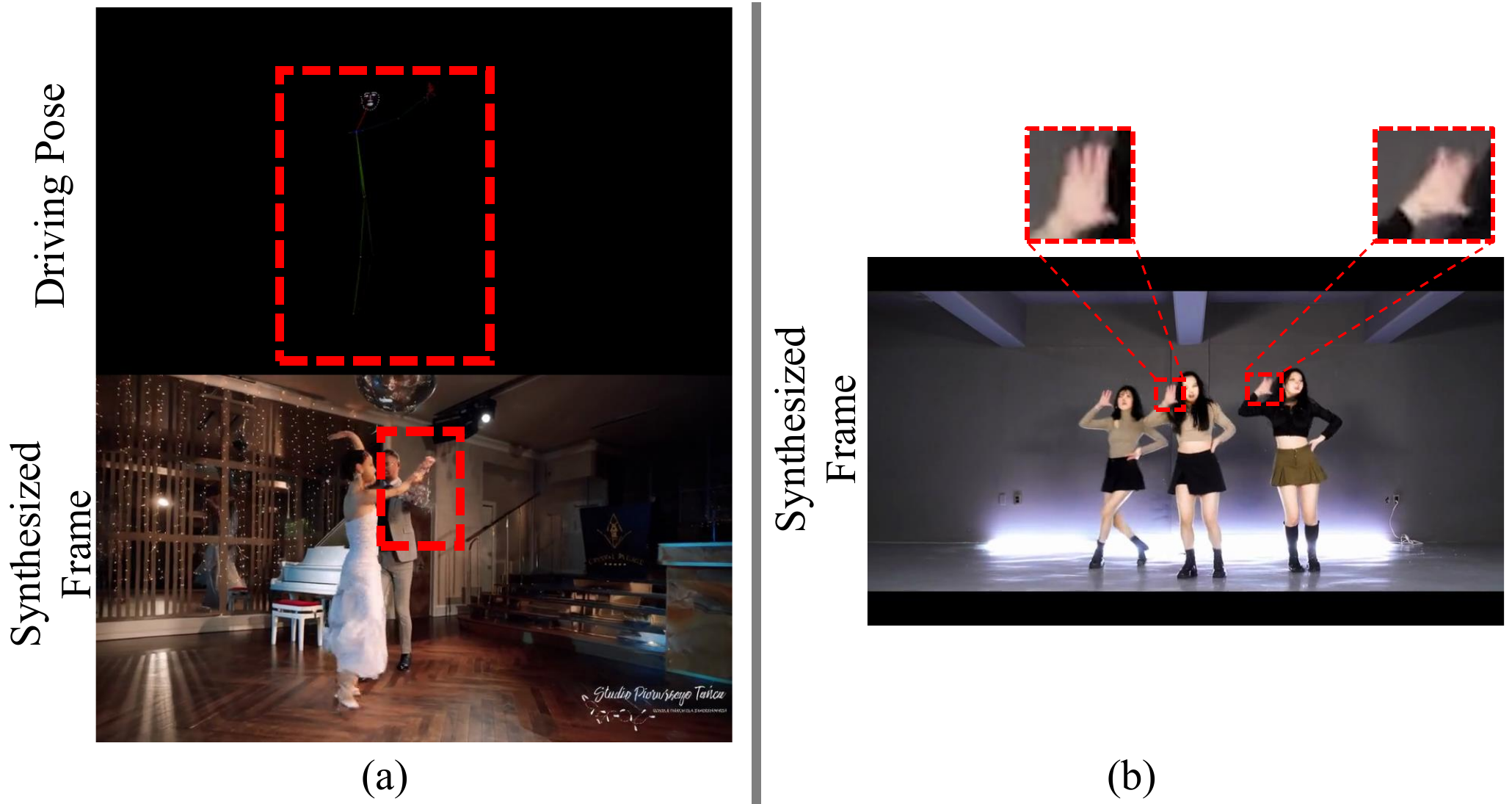}
  \caption{Limitation Analysis. (a) The model produces artifacts when the input pose sequence is extremely noisy or incomplete. (b) Fine-grained anatomical structures, such as hands, are prone to artifacts due to the 480p resolution training bottleneck.}
  \label{fig:limitation}
\end{figure}

\section{More Qualitative Results}
\label{sec:results}

In this section, we provide additional qualitative results, including extended comparisons with baselines, various application examples, and our strategy for long-duration video generation. To better evaluate motion smoothness and identity stability, we provide video counterparts for all qualitative figures from both the main paper and this appendix. These videos are available on the project webpage, where long-duration generation results are displayed at 15 FPS, while all other comparisons and application examples are presented at 10 FPS.

\noindent\textbf{More Qualitative Comparisons with SOTAs.}
To further demonstrate the superiority of AnyCrowd, we provide visual comparisons with state-of-the-art baselines in Fig.~\ref{fig:selfdriven_comp} and Fig.~\ref{fig:crossdriven_comp}, covering \emph{self-driven} and \emph{cross-driven} scenarios, respectively. In the self-driven setting, AnyCrowd exhibits more precise identity-to-pose correspondence, effectively mitigating common artifacts such as identity swapping or blending observed in competing methods. For the more challenging cross-driven setting, our model significantly outperforms existing SOTAs in maintaining temporal identity consistency and structural integrity throughout the entire sequence.

\noindent\textbf{More Application Results.} 
Fig.~\ref{fig:applications} showcases the versatility of AnyCrowd across various application scenarios: (a) \textbf{flexible identity inputs} from diverse sources; (b) \textbf{one-to-many animation}, where a single reference character is driven by multiple distinct pose sequences; (c) \textbf{many-to-one animation}, illustrating the synchronization of multiple character identities with a single driving pose sequence; and (d) \textbf{arbitrary pose-id mapping}, demonstrating the model's precise control in reassigning character identities to specific target poses within a scene.

\noindent\textbf{Long Video Generation.}
To generate videos that exceed the duration seen during training, we adopt a sliding-window denoising strategy, as utilized in prior works such as UniAnimate-DiT~\cite{wang2025unianimate}. Specifically, we partition the entire latent token sequence into multiple overlapping segments. During each denoising step, these segments are processed independently by the model. To ensure temporal consistency and seamless transitions across segment boundaries, we maintain a global accumulation buffer that tracks the update frequency for each latent position. The final latent representation at each timestep is obtained by aggregating the denoised outputs from all overlapping segments, normalized by their respective update counts. This effectively performs a weighted temporal blending across segments.
With this sliding-window mechanism, our AnyCrowd—though trained on 49-frame sequences—can successfully generate 98-frame results. For visual results of long video generation, please refer to the end of the project webpage.

\section{Limitations and Broader Impacts}
\label{sec:limitation}

\noindent\textbf{Limitations.} 
Despite the effectiveness of AnyCrowd, we acknowledge two primary limitations that warrant further investigation. First, the generation quality is generally robust to moderate pose variations, yet the model may produce artifacts when faced with extremely noisy, incomplete, or anatomically implausible 2D pose sequences. As shown in Fig.~\ref{fig:limitation}(a), this sensitivity to extreme outliers stems from the model's primary training on high-quality pose data. To mitigate this, we plan to incorporate robust data augmentation strategies, such as applying stochastic perturbations to 2D poses during training, and integrating textual guidance via Vision-Language Models (VLMs) to provide semantic supervision that complements pose control. Second, our model struggles with synthesizing fine-grained details, particularly in human hands (as shown in Fig.~\ref{fig:limitation}(b)), which is largely attributable to the resolution bottleneck of our 480p training framework. Scaling the training process to higher resolutions, such as 720p, is a promising path to enhance the model's capacity to preserve intricate anatomical details.

\noindent\textbf{Broader Impacts.}
AnyCrowd focuses on multi-character animation for arbitrary numbers of characters, a task of significant research value given the inherent complexity of modeling multi-agent interactions. Our work holds substantial promise across diverse domains, including the film industry for efficient production, virtual reality (VR) for immersive content creation, and synthetic data generation for embodied AI, where scalable and controllable character animation is increasingly in demand. Furthermore, the proposed identity-isolated representation and id-pose binding mechanism offer valuable architectural insights for multi-subject, multi-modal video generation. We believe these contributions provide a solid foundation for more generalizable generative models, thereby fostering broader advancements in the field of generative AI.

\end{document}